\documentclass{article}

\usepackage[final]{corl_2026}

\usepackage{amsmath}
\usepackage{amssymb}
\usepackage{algorithm}
\usepackage{algpseudocode}
\usepackage{xcolor}
\usepackage{float}
\usepackage{graphicx}
\usepackage{booktabs}
\usepackage{mathtools}
\usepackage{booktabs}
\usepackage{multirow}
\usepackage{wrapfig}
\usepackage{enumitem}
\usepackage{caption}
\usepackage{subcaption}
\usepackage{tabularx}

\title{Learning Decomposed Visibility for Efficient Active Exploration of Cluttered Scenes}

\author{
  Suyun Lee$^{*}$ \quad Minsoo Choi$^{*}$ \quad Jihwan Gong \quad Unghui Nam \quad Minji Bae \quad Byonghyo Shim$^{\dagger}$\\[4pt]
  Seoul National University\\
  \texttt{\{sylee, mschoi, uhnam, mjbae, bshim\}@islab.snu.ac.kr}\\[6pt]
  {\small $^{*}$Equal Contribution \quad $^{\dagger}$Corresponding Author}
}
\begin{document}
\maketitle

%===============================================================================

\begin{abstract}
Active exploration in cluttered environments is essential for reconstructing occluded 3D scenes, requiring both observation and physical interaction. Existing methods rely on belief maps that indicate where the scene is unresolved but not how each region can be resolved, estimating the gains of viewing and pushing only through costly lookahead or hand-tuned heuristics.
We propose DiVE (Decomposed Visibility Estimator), a learned model that predicts two decomposed visibility maps: a view-resolvable map and a push-resolvable map, making explicit what each action mode can reveal.
By converting these modality-specific predictions into a common visibility-gain scale, DiVE supports a hierarchical decision structure: the robot first chooses between observation and interaction, then selects the specific action through a mode-specific policy.
In simulation, DiVE improves occupancy IoU by $11.3\%$ on extreme-occlusion shelf scenes and runs $13.2\times$ faster per step ($19.8\times$ faster in planning) than a state-of-the-art baseline.
Deployed zero-shot on a real robot, it recovers $79.2\%$ of initially occluded objects.
\textbf{Project page:} \url{https://lee-su-yun.github.io/DiVE/}
\end{abstract}


\section{Introduction}
\vspace{-5pt}

Recent advances in robot learning have significantly expanded the range of tasks that robots can perform in real-world environments, from general-purpose manipulation to mobile and long-horizon interaction~\citep{brohan2022rt, zitkovich2023rt, kim2024openvla, black2024pi_0, fu2024mobile, liu2024ok}.
As robots are deployed in increasingly complex environments, successful action depends not only on control but also on an accurate understanding of the scene in which the action takes place.
Successful task execution therefore requires the robot to recover the geometric and semantic structure of the scene, including object locations and categories~\citep{rosinol2020kimera, xie2021unseen}.
For example, to retrieve a mug hidden behind other objects on a kitchen shelf, a robot must understand the arrangement of nearby objects and the available free space to determine how to access the mug.

In particular, scene understanding is challenging in cluttered environments because task-relevant objects and the free space are often occluded from the robot’s current view. The robot must therefore actively explore the scene through viewpoint changes and physical interactions: a viewpoint change can reveal regions hidden from the current camera pose, whereas physical interaction can expose regions that remain hidden from all accessible viewpoints. For effective exploration, the robot must coordinate these two modes, deciding at each step when to acquire a new view and when to interact with the scene.

\begin{figure}[t]
    \centering
    \includegraphics[width=\linewidth]{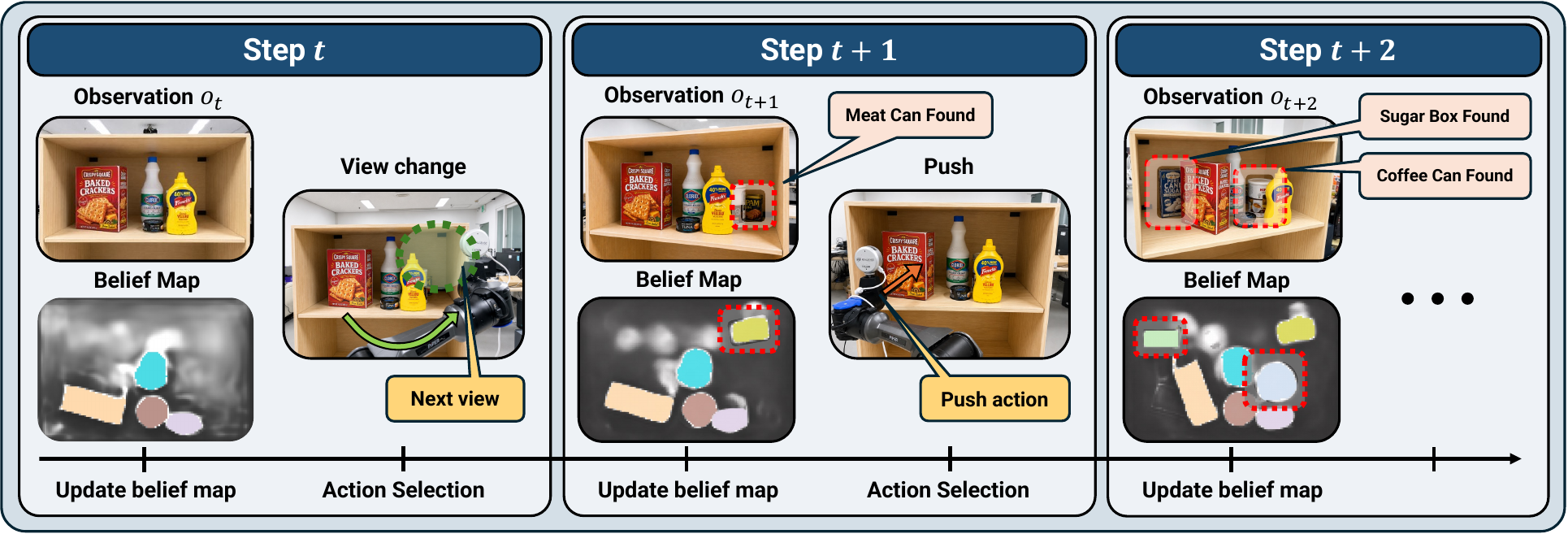}
    \vspace{-10pt}
    \caption{\textbf{Example scenario in a real shelf environment.} At each step, the belief map estimator updates the global scene belief from the current observation, visualized here as a 2D top-down projection. Given the visibility maps, the planner decides whether to change the viewpoint or execute a push. In this example, the planner selects a view change at step $t$ and executes a push at step $t+1$.}
    
    \label{fig:example}
    \vspace{-10pt}
\end{figure}

Recent methods address this problem by jointly planning viewpoint changes and physical interactions~\citep{Marques2025, Dengler2023}. They typically rely on a shared belief or uncertainty map to represent what remains hidden in the scene.
While such a representation indicates where the scene remains unresolved, it does not specify how each hidden region can be resolved---by moving the camera or by physically changing the scene. 
As a result, the expected gain of viewing versus pushing is not directly available from the representation itself. 
Existing methods estimate this quantity indirectly, either by precomputing which regions each candidate action would reveal and scoring each action against the current uncertainty, or by relying on hand-tuned switching thresholds. 
In both cases, action selection is imposed on top of the representation rather than derived from it, resulting in either substantial computational cost or limited adaptability to changing scene geometry and occlusion patterns.

In this work, we focus on visibility as the basis for scene representation, since it is directly affected by viewpoint changes and physical interactions. Rather than representing only where uncertainty remains, our representation characterizes the resolvability of hidden regions under different action modalities. The key feature is that visibility provides an action-resolvable signal: some regions are occluded only from the current view and can be revealed by moving the camera, whereas others remain hidden from all accessible viewpoints until physical interaction changes the scene geometry.

Building on this insight, we propose the \textbf{Decomposed Visibility Estimator (DiVE)}, a learned model that predicts modality-specific visibility from partial observations.
Given the partial observation, DiVE estimates two complementary maps: 1) a view-resolvable visibility map $\phi_{\text{view}}$, which captures regions likely to become visible through viewpoint changes; and 2) a push-resolvable visibility map $\phi_{\text{push}}$, which captures regions likely to become visible through pushing.
By separating these two forms of visibility, DiVE directly encodes the action-specific resolvability of hidden regions, rather than recovering it indirectly from a shared uncertainty map through external scoring rules.

We adapt the established hierarchical planning structure to our decomposed representation by deriving a common criterion based on the expectation of visibility change. This criterion maps the differently encoded modality-specific visibility signals onto a shared measure, enabling direct comparison between viewing and pushing.
At each step, the planner converts the modality-specific maps to a common expected-visibility-change scale, selects between viewing and pushing, and then invokes a mode-specific policy to rank the corresponding candidates.
The selected policy ranks the candidates in a single forward pass, avoiding candidate-wise post-action belief prediction.

In summary, our contributions are as follows:
\vspace{-4pt}
\begin{itemize}[topsep=2pt, itemsep=4pt, parsep=0pt]
    \item We introduce decomposed visibility, an action-aware representation that distinguishes regions resolvable by viewpoint changes from those resolvable through physical interaction. We realize this representation with the Decomposed Visibility Estimator (DiVE), which predicts view-resolvable and push-resolvable visibility maps from partial observations. By preserving which action modality can resolve each region, the representation exposes action-relevant structure and directly informs whether additional sensing or physical interaction is more informative.

    \item We adapt the established hierarchical view-and-push planning structure to the decomposed visibility representation.
    Accordingly, we derive a common expected-visibility-change criterion that makes the modality-specific signals directly comparable without hand-tuned cross-mode balancing.
    We further use mode-specific policies for lightweight within-mode candidate ranking, replacing per-candidate belief lookahead with a single forward pass.

    \item We validate our framework in both simulation and real-world cluttered shelf scenes.
    In simulation, DiVE improves occupancy IoU by $11.3\%$ under extreme occlusion and runs $13.2\times$ faster per step than a state-of-the-art baseline.
    On a real robot, DiVE transfers zero-shot and recovers $79.2\%$ of initially occluded objects.
\end{itemize}
\section{Related Work}
\vspace{-6pt}

\subsection{Next-Best-View Planning}
\vspace{-6pt}
Next-best-view (NBV) planning selects viewpoints by maximizing information gain over a volumetric scene representation~\citep{connolly1985determination, Isler2016, Delmerico2018, daudelin2017adaptable, bircher2016receding}.
Later work reduces ray-casting cost~\citep{PBNBV2025} or replaces hand-crafted criteria with learned viewpoint prediction~\citep{GenNBV2024, peralta2020next, jin2023neu}.
Even when extended to manipulator-mounted cameras in confined clutter~\citep{Ren2023}, these methods treat the scene as static and assume hidden geometry can be recovered from viewpoints alone~\citep{feng2024naruto, jiang2024fisherrf, xue2024neural}, either by direct observation or by shape inference.
This assumption fails in dense clutter, where some regions remain occluded from every accessible viewpoint unless the scene is physically altered.

\vspace{-6pt}
\subsection{Interactive Perception}
\vspace{-6pt}
Interactive perception~\citep{Bohg2017, KatzBrock2008} extends active perception~\citep{bajcsy1988active, bajcsy2018revisiting, chen2011active} by allowing the robot to physically alter the scene rather than only sampling it from different viewpoints.
Physical interaction has improved perception in tasks such as instance segmentation in dense clutter~\citep{eitel2019self, Serhan2022, RISeg2024} and retrieving target objects occluded by surrounding items~\citep{Danielczuk2019, kurenkov2020visuomotor, danielczuk2020x, huang2021visual}.
These methods use manipulation to reveal hidden information, generally with the camera held fixed.
This fixed-camera assumption motivates planning viewpoint changes and physical interactions within a single decision process.

\vspace{-6pt}
\subsection{Combined Viewpoint and Manipulation Planning}
\vspace{-6pt}
Recent work jointly plans viewpoints and pushes, recognizing that neither modality alone suffices in clutter.
Early approaches use reinforcement learning to learn policies that switch between viewing and pushing for object search~\citep{Novkovic2020, zenkri2022hierarchical}.
On cluttered shelves, \citet{Dengler2023} trigger a push only after viewpoint information gain saturates, using hand-tuned thresholds; a follow-up work replaces this heuristic with uncertainty-informed selection~\citep{Dengler2025}.
\citet{Marques2025} formulate the problem as a POMDP in metric-semantic map space and solve it with receding-horizon planning using calibrated neural belief updates.
Despite these advances, existing methods do not provide a representation-level gain signal that makes viewing and pushing directly comparable.
Consequently, action selection still relies on expensive lookahead~\citep{Marques2025} or heuristic mode switching~\citep{Dengler2023}.
DiVE addresses this limitation by decomposing visibility into modality-specific signals that enable direct comparison of view and push gains. For within-mode action selection, DiVE uses lightweight distilled policies to rank candidate actions in a single forward pass.

\section{Problem Definition}
\vspace{-9pt}
We consider the problem of active scene reconstruction in cluttered shelf environments, where a robot builds a geometric and semantic representation of the scene from partial observations.
Due to the shelf's spatial constraints, we consider two complementary action types: viewpoint selection for sensing around occlusions and pushing for altering occlusion geometry.
The action space is $\mathcal{A}=\mathcal{C}\cup\mathcal{P}$, where $\mathcal{C}$ contains candidate viewpoints and $\mathcal{P}$ contains candidate pushes parameterized by contact point and direction.
The scene is represented as a voxel grid $\mathcal{V}$ with labels $z\in\{0,1,\dots,N_{\text{classes}}\}$~\citep{moravec1985high, curless1996volumetric}, where 0 denotes free space and the remaining labels denote object categories.
At each time step $t$, let $s_t$ denote the current scene state and $\phi_{\text{scene},t}$ the robot's belief about that state. The objective is to reconstruct the scene accurately while selecting informative actions efficiently.
\section{Methodology}
\vspace{-6pt}
\begin{figure}[t]
    \centering
    \includegraphics[width=\linewidth]{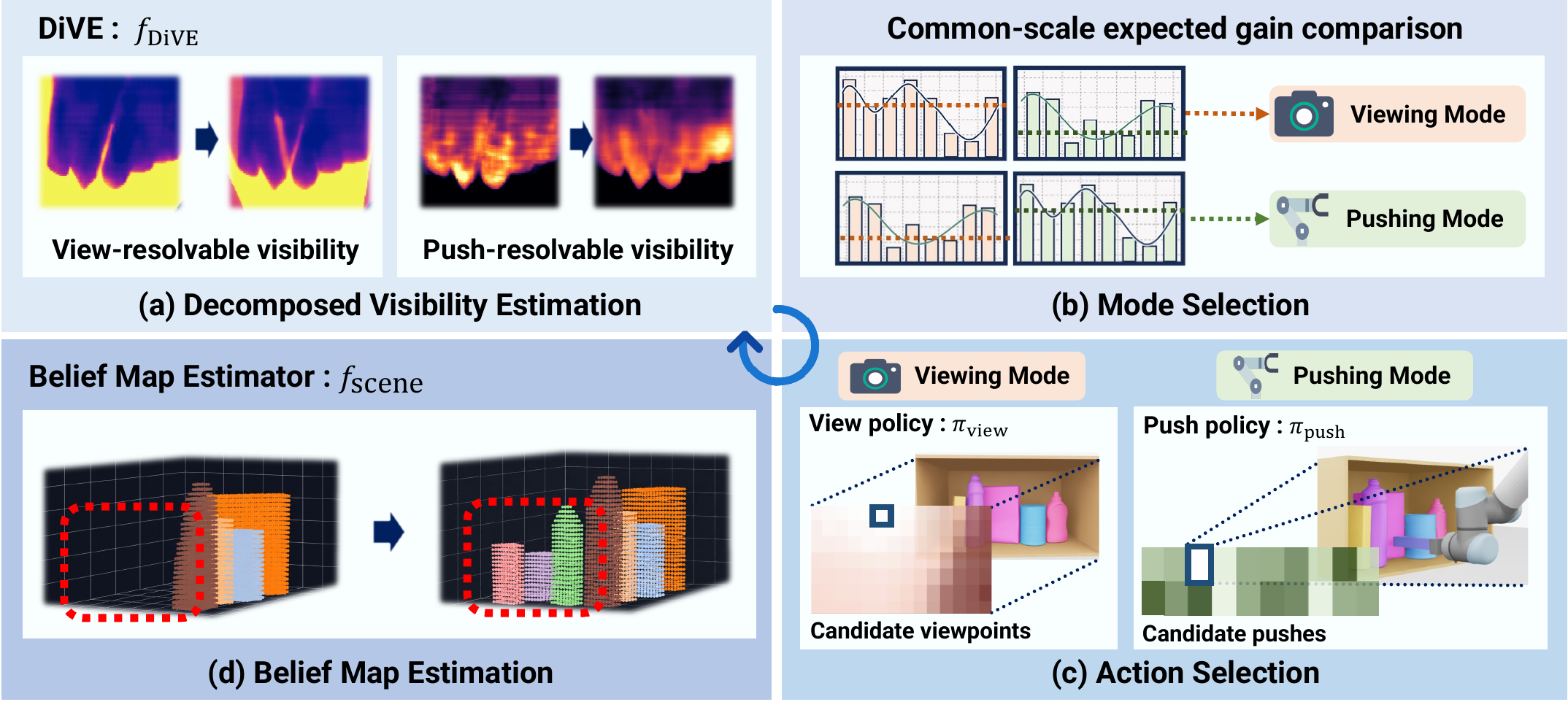}
    \vspace{-15pt}
    \caption{\textbf{Overview of the proposed framework.}
    \textbf{(a)} DiVE ($f_{\text{DiVE}}$) predicts view-resolvable ($\phi_{\text{view}}$) and push-resolvable ($\phi_{\text{push}}$) visibility maps from the current partial observation.
    \textbf{(b)} The maps are converted to a common scale of expected visibility change to select the action mode.
    \textbf{(c)} The corresponding policy ($\pi_{\text{view}}$ or $\pi_{\text{push}}$) selects the best viewpoint or push.
    \textbf{(d)} The belief map estimator ($f_{\text{scene}}$) fuses the new observation and, after a push, the swept region to update the global scene belief $\phi_{\text{scene}}$.}

    \label{fig:overview}
\end{figure}

\subsection{Overview}
\vspace{-6pt}
Our method separates two questions that are often coupled in active scene reconstruction:
\textit{how should the robot maintain a global scene estimate from partial observations?} and
\textit{which action can make hidden regions visible?}
We address the first with a belief map estimator that integrates observations into a global 3D scene belief, and the second with decomposed visibility maps that drive hierarchical view-and-push action selection.
As shown in Fig.~\ref{fig:overview}, our framework consists of three components:
DiVE, mode-and-action selection, and a belief map estimator. 
\vspace{-6pt}
\subsection{Decomposed Visibility Estimator (DiVE)}
\label{sec:DiVE}
\vspace{-6pt}
The role of DiVE is to convert a partial observation into an action-resolvable visibility representation. 
A unified map of unresolved regions does not specify whether each region is resolvable by viewing or by physical interaction.
We represent visibility with two modality-specific maps: $\phi_{\text{view}}$, which captures regions whose visibility depends on viewpoint selection, and $\phi_{\text{push}}$, which captures regions whose visibility can change after pushing.
This decomposition preserves the action modality responsible for resolving each region, providing the signal needed for comparing viewing and pushing in Sec.~\ref{sec:mode_action_select}.

At each time step $t$, DiVE takes the current observation $o_t$ and the previous visibility maps as input, and predicts the updated maps:
\begin{equation}
(\phi_{\text{view},t},\ \phi_{\text{push},t}) 
= f_{\text{DiVE}}(\phi_{\text{view},t-1},\ \phi_{\text{push},t-1},\ o_t).
\end{equation}
The two output heads share the same architecture, but each is trained with a different target to capture how viewpoint changes and pushing affect visibility.
\vspace{-5pt}

% --- View head ---
\paragraph{View-resolvable visibility ($\phi_{\text{view}}$).}
The view head estimates which regions can be resolved through additional viewpoints.
For each voxel $v \in \mathcal{V}$, it models how often that voxel is visible from the viewpoints visited since the most recent push.
Let $m_k \in \{0,1\}^{|\mathcal{V}|}$ denote the binary visibility map obtained from camera $k$ on the current scene, and let $\mathcal{K}_t$ denote the set of viewpoints visited since the most recent push, or since the episode start if no push has occurred.
We define the training target as the empirical visibility frequency over $\mathcal{K}_t$:
\begin{equation}
\phi^*_{\text{view},t}(v) 
= \frac{1}{|\mathcal{K}_t|} \sum_{k \in \mathcal{K}_t} m_k(v).
\label{eq:gt_view}
\end{equation}
Intuitively, $\phi_{\text{view}}$ represents how frequently each voxel is visible across visited viewpoints: high values correspond to voxels visible from many views, and low values to voxels rarely visible from the current view set. 
Thus, $\phi_{\text{view}}$ serves as a visibility state, rather than a direct gain map, from which the expected gain of an additional viewpoint is derived in Sec.~\ref{sec:mode_action_select}.
\vspace{-5pt}
% --- Push head ---
\paragraph{Push-resolvable visibility ($\phi_{\text{push}}$).}
The push head estimates which regions can be resolved through physical interaction.
Unlike viewpoint changes, pushing alters the scene geometry; a useful push is therefore one that changes the marginal visibility of voxels across the candidate viewpoint set.
For a push $p$, let 
$\bar{m}^{\text{before}}(v) = \mathbb{E}_{k \sim \mathcal{C}}[m_k^{\text{before}}(v)]$ 
and 
$\bar{m}^{\text{after}}_p(v) = \mathbb{E}_{k \sim \mathcal{C}}[m_{k,p}^{\text{after}}(v)]$ 
denote the marginal visibility of voxel $v$ over candidate viewpoints before and after executing $p$, respectively.
We define the training target as the expected absolute change in marginal visibility induced by a randomly sampled candidate push:
\begin{equation}
\phi^*_{\text{push},t}(v) 
= \mathbb{E}_{p \sim \mathcal{P}_t}
\Big[\,\big|\bar{m}^{\text{after}}_p(v) - \bar{m}^{\text{before}}(v)\big|\,\Big],
\label{eq:gt_push}
\end{equation}
where $\mathcal{P}_t$ and $\mathcal{C}$ denote the candidate push and viewpoint sets, respectively.
For supervision generation, every push candidate is executed from the same restored pre-push configuration, and all 98 candidate viewpoints are re-rendered after execution; full details are provided in the Appendix.
Unlike $\phi_{\text{view}}$, which represents a visibility state, $\phi_{\text{push}}$ directly represents an expected visibility change induced by pushing.
This target assigns high values to voxels whose visibility is likely to change after pushing, capturing regions that are not merely unobserved but physically resolvable through interaction.

\vspace{-5pt}
\subsection{Mode and Action Selection}
\label{sec:mode_action_select}
\vspace{-5pt}
Given the decomposed visibility maps from $f_{\text{DiVE}}$, we select an action in two stages.
First, we select the action mode by comparing the expected visibility gains of viewing and pushing.
Then, conditioned on the selected mode, a mode-specific policy selects the best candidate viewpoint or push.
This hierarchical design avoids directly ranking heterogeneous view and push candidates against one another, which can be unstable because their predicted gains arise from different action effects.
By comparing modes before ranking candidates, the planner becomes less sensitive to noisy or outlier candidate scores.
\vspace{-5pt}
% --- Mode selection ---
\paragraph{Mode selection.}
The main challenge in mode selection is that the two visibility maps have different meanings.
The view map $\phi_{\text{view},t}$ is a per-voxel visibility state accumulated over viewpoints, whereas the push map $\phi_{\text{push},t}$ already represents an expected visibility change induced by pushing.
To compare viewing and pushing, we convert both quantities into the same unit: the expected per-voxel visibility change induced by one action of the corresponding mode.
We use the absolute change in visibility to measure how much an action changes the visibility state, regardless of whether the change corresponds to newly revealed or newly occluded regions.

For viewing, $\phi_{\text{view},t}$ is a running average over the viewpoints visited so far.
Thus, the effect of one additional viewpoint can be measured by how much the updated visibility state $\phi_{\text{view},t+1}$ differs from $\phi_{\text{view},t}$.
Since the true distribution of visibility over unvisited viewpoints is not available at test time, we use $\phi_{\text{view},t}(v)$ as a plug-in marginal estimate of the Bernoulli mean for the binary visibility $m_{k_{t+1}}(v)$ observed from a newly selected viewpoint $k_{t+1}\sim\mathcal{C}$.
Under this plug-in model, the expected absolute update for voxel $v$ is
\begin{equation}
\mathbb{E}_{k_{t+1} \sim \mathcal{C}}\bigl[\,|\phi_{\text{view},t+1}(v) - \phi_{\text{view},t}(v)|\,\bigr]
= \frac{2\,\phi_{\text{view},t}(v)\,(1 - \phi_{\text{view},t}(v))}{|\mathcal{K}_t|+1},
\label{eq:view_change}
\end{equation}
with the derivation and empirical validation provided in the Appendix.
Empirically, the approximation shows strong agreement after a few accumulated viewpoints, with normalized mean absolute error (nMAE) below $10\%$ for $|\mathcal{K}_t|\geq5$ across both low- and high-occlusion scenes.
We use this quantity as a mode-selection proxy: it is largest when the current visibility state is uncertain across viewpoints and decreases as a voxel becomes consistently visible or consistently hidden.

The push map $\phi_{\text{push},t}$ is already trained to predict the expected absolute change in marginal visibility caused by pushing.
Therefore, after Eq.~\ref{eq:view_change}, both viewing and pushing are expressed as expected per-voxel visibility changes.
The resulting mode-level gains are
\begin{equation}
\mathcal{G}_\text{view}(\phi_{\text{view},t})
= \sum_{v \in \mathcal{V}}
\frac{2\,\phi_{\text{view},t}(v)\,(1 - \phi_{\text{view},t}(v))}{|\mathcal{K}_t|+1},
\quad
\mathcal{G}_\text{push}(\phi_{\text{push},t})
= \sum_{v \in \mathcal{V}} \phi_{\text{push},t}(v).
\label{eq:gain}
\end{equation}
The mode with the higher expected gain is selected as
$\mathcal{M}_t = \operatorname*{arg\,max}_{m \in \{\text{view}, \text{push}\}} \mathcal{G}_m$. Because both gains measure the expected visibility change from a single action, the mode decision is derived directly from the decomposed visibility maps rather than from a hand-tuned switching rule.
To avoid applying this comparison to a severely under-observed visibility state, we require at least two accumulated views at the start of an episode and after each push; until then, the planner selects the viewing mode.

\vspace{-5pt}

 % --- Action selection ---
\paragraph{Action selection.}
Once the mode $\mathcal{M}_t$ is selected, we choose the specific action within the selected mode using the corresponding policy, $\pi_{\text{view}}$ or $\pi_{\text{push}}$.
These policies replace online one-step lookahead at test time with learned candidate-ranking functions~\citep{frahm2025vin, hinton2015distilling}.
Since the mode has already been selected, each policy only needs to compare candidates within a homogeneous action set, making the ranking problem simpler and avoiding cross-mode score calibration.

The view selection policy $\pi_{\text{view}}$ takes the current view-resolvable visibility map $\phi_{\text{view},t}$ and the set of visited viewpoints $\mathcal{K}_t$ as input, and predicts a gain score for each candidate viewpoint $k \in \mathcal{C}$:
\begin{equation}
\pi_{\text{view}}: (\phi_{\text{view},t}, \mathcal{K}_t) 
\mapsto \{\hat{G}_{\text{view}}(k)\}_{k \in \mathcal{C}},
\quad
\hat{G}_{\text{view}}(k) \approx 
\frac{g_{\text{view}}(k)-\mu_{\text{view}}}{\sigma_{\text{view}}},
\end{equation}
where
$
g_{\text{view}}(k)
=
\sum_{v \in \mathcal{V}}
\left|
\phi_{\text{view},t+1}(v;\,k)
-
\phi_{\text{view},t}(v)
\right|
$ is the one-step visibility gain of viewpoint $k$, obtained by updating $\phi_{\text{view}}$ with the pretrained view head of $f_{\text{DiVE}}$ after observing from viewpoint $k$.
The mean $\mu_{\text{view}}$ and standard deviation $\sigma_{\text{view}}$ are computed over all candidate viewpoints in $\mathcal{C}$.
This normalization makes the policy predict relative candidate quality within the current viewpoint set, rather than requiring calibrated absolute gain values.

The push selection policy $\pi_{\text{push}}$ similarly ranks candidate pushes within $\mathcal{P}_t$.
Each push candidate $p$ is represented by its swept volume, which indicates the region directly affected by the interaction.
Given the push-resolvable visibility map $\phi_{\text{push},t}$ and the candidate push set $\mathcal{P}_t$, the policy predicts
\begin{equation}
\pi_{\text{push}}: (\phi_{\text{push},t}, \mathcal{P}_t)
\mapsto \{\hat{G}_{\text{push}}(p)\}_{p \in \mathcal{P}_t},
\quad
\hat{G}_{\text{push}}(p) \approx 
\frac{g_{\text{push}}(p)-\mu_{\text{push}}}{\sigma_{\text{push}}},
\end{equation}
where
$
g_{\text{push}}(p)
=\sum_{v \in \mathcal{V}}\left|\bar{m}^{\text{after}}_p(v)-\bar{m}^{\text{before}}(v)\right|$
is the one-step visibility gain of push $p$.
Here, $\bar{m}^{\text{before}}$ and $\bar{m}^{\text{after}}_p$ denote the marginal visibility maps before and after executing $p$, as defined in Sec.~\ref{sec:DiVE}.
The mean $\mu_{\text{push}}$ and standard deviation $\sigma_{\text{push}}$ are computed over the candidate push set $\mathcal{P}_t$.

The next action is selected by maximizing the predicted gain score within the selected mode.
At test time, the policies select actions with a single forward pass over the candidate set, avoiding repeated simulation or visibility recomputation for every candidate.
\vspace{-5pt}
\subsection{Belief Map Estimator}
\label{sec:beliefmap}
\vspace{-5pt}
While DiVE guides action selection, the belief map estimator maintains the geometric and semantic reconstruction.
For geometry, each voxel occupancy is represented by a Beta distribution parameterized by $(\alpha_t,\beta_t)$, whose mean $\alpha_t/(\alpha_t+\beta_t)$ gives a bounded occupancy estimate while separating occupied and free-space evidence~\citep{Marques2025, Dengler2025, asgharivaskasi2021active, matez2024voxeland}.
For semantics, $z_t$ denotes a per-class voxel field, and the scene belief is $\phi_{\text{scene},t}=(\alpha_t,\beta_t,z_t)$.
At each time step, $f_{\text{scene}}$ updates this belief from $\phi_{\text{scene},t-1}$, the current observation $o_t$, and, when applicable, the swept volume $\Omega_t$:
\begin{equation}
\phi_{\text{scene},t}
=
f_{\text{scene}}\bigl(\phi_{\text{scene},t-1},\, o_t,\, \Omega_t\bigr).
\end{equation}
By separating scene belief from decomposed visibility, the framework uses visibility for efficient action selection while retaining a dedicated representation for reconstruction.

\vspace{-5pt}

% 내용 작성
\begin{figure}[t]
\centering
% ── 왼쪽: (a) main result ──────────────────────────
\begin{subfigure}[t]{0.70\textwidth}
    \vspace{0pt}
    \centering
    \includegraphics[width=\linewidth]{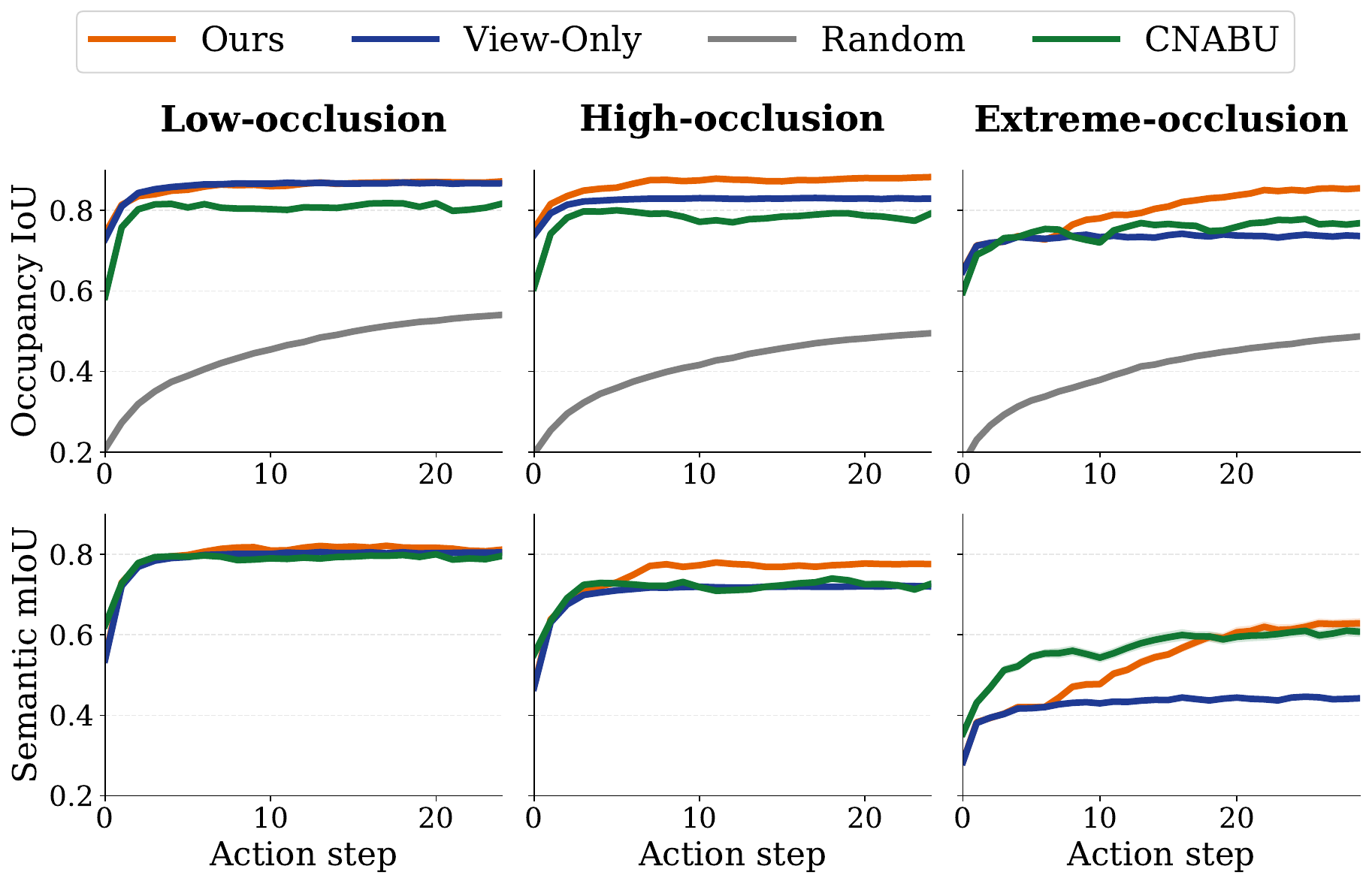}
    \vspace{-15pt} 
    \caption{}
    \label{fig:main_result}
\end{subfigure}\hfill
% ── 오른쪽: (b) OOD 위 + (c) table 아래 ────────────
\begin{minipage}[t]{0.27\textwidth}
    \vspace{0pt}
    % --- (b) OOD graph ---
    \begin{subfigure}[t]{\linewidth}
        \vspace{0pt}
        \centering
        \includegraphics[width=\linewidth]{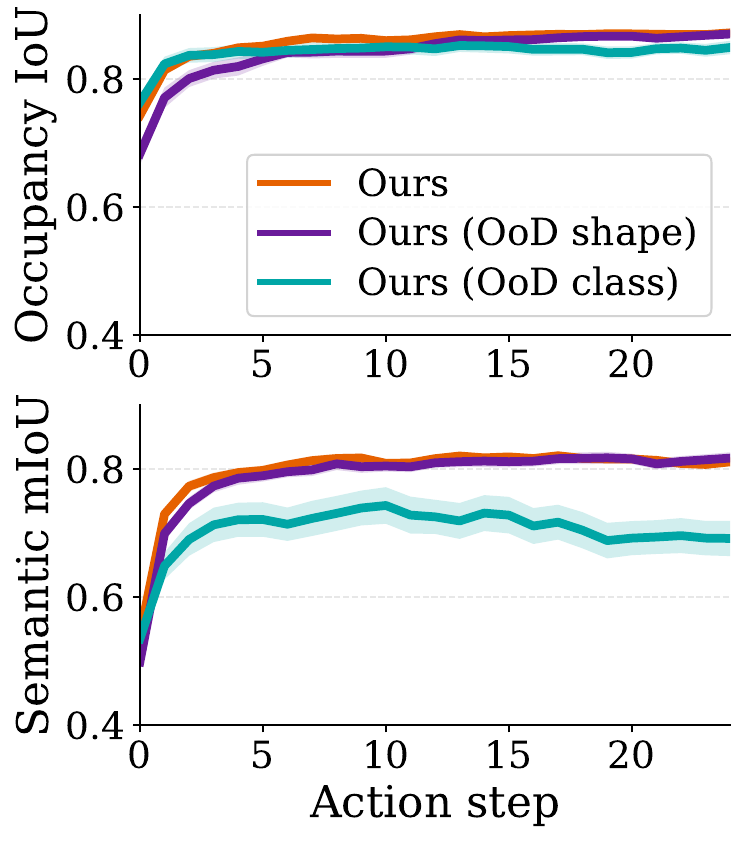}
        \vspace{-17pt}        % ← (b) 라벨을 그림 쪽으로 올림
        \caption{}
        \label{fig:ood}
    \end{subfigure}
    \vspace{0pt}
   % --- (c) efficiency table ---
    \begin{subtable}{\linewidth}
        \vspace{5pt} 
        \centering
        \setlength{\tabcolsep}{3pt}
        \renewcommand{\arraystretch}{0.5} 
        {\fontsize{7}{8.4}\selectfont
        \begin{tabular}{lcc}
            \toprule
            Method & Planning & Total \\
            \midrule
            CNABU           & 9.14 $\pm$ 1.29 & 9.31 $\pm$ 1.31 \\
            \textbf{Ours}  & \textbf{0.46 $\pm$ 0.04} & \textbf{0.70 $\pm$ 0.07} \\
            \midrule
            Speedup               & 19.8$\times$ & 13.2$\times$ \\
            \bottomrule
        \end{tabular}
        }
        \vspace{-4pt}        % ← (c) 라벨을 표 쪽으로 올림
        \caption{}
        \label{tab:efficiency}
    \end{subtable}
\end{minipage}
\vspace{-5pt} 
% ── 전체를 설명하는 캡션 하나 ──────────────────────
% \caption{\textbf{Main results, generalization, and efficiency.}
% \textbf{(a)} Reconstruction quality vs.\ action budget on low-, high-, and extreme-occlusion scenes (occupancy IoU, top; semantic mIoU, bottom), comparing DiVE against CNABU~\citep{Marques2025}, \emph{View-Only}, and \emph{Random}.
% \textbf{(b)} Generalization to objects with out-of-distribution shapes or classes; performance degrades only marginally.
% \textbf{(c)} Per-step compute over $100$ episodes. \emph{Planning} is the time to select the next action (viewpoint or push); \emph{Total} adds belief update and evaluation.}
\caption{\textbf{Main results, generalization, and efficiency.}
\textbf{(a)} Reconstruction quality vs.\ action budget on low-, high-, and extreme-occlusion scenes, comparing DiVE against CNABU~\citep{Marques2025}, \emph{View-Only}, and \emph{Random}.
\textbf{(b)} Generalization to objects with out-of-distribution shapes or classes.
\textbf{(c)} Per-step compute over $100$ episodes. \emph{Planning} is the time to select the next action; \emph{Total} additionally includes belief update and evaluation.} \vspace{-6pt}
\label{fig:results}
 
\end{figure}

\section{Experiments}
\vspace{-5pt}
\subsection{Experimental Setup}
\label{sec:setup}
\vspace{-5pt}
We evaluate DiVE in two settings: \emph{simulation}, using a UR5 in Isaac Sim~\citep{isaacsim} in front of a %$60{\times}30{\times}40$~cm 
cluttered shelf with YCB objects~\citep{calli2015ycb}, and \emph{real-world}, using a Piper AgileX with a comparable shelf and objects from the same categories, transferred zero-shot.
We test on three levels of occlusion---\emph{low}, \emph{high}, and \emph{extreme}---with the model trained only on low- and high-occlusion scenes; extreme occlusion is held out for evaluation. 
Full details on scenes, data, training, and metrics are provided in the Appendix.
%Full details on scenes, data, training, and metrics are in Appendix~\ref{app:implementation}--\ref{app:experimental_details}.
\vspace{-5pt}

% \begin{wrapfigure}{r}{0.6\textwidth}
%     \centering
%     \vspace{-12pt}
%     \includegraphics[width=0.58\textwidth]{_CoRL_2026__DiVE/figures/main_result.pdf}
%     \vspace{-8pt}
%     \caption{\textbf{Reconstruction quality versus action budget} on low-, high-, and extreme-occlusion shelf scenes, measured by occupancy IoU (top) and semantic mIoU (bottom).
% We compare DiVE against CNABU~\citep{Marques2025}, a manipulation-aware planner that jointly selects view and push actions; \emph{View-Only}, an ablation of DiVE restricted to $\pi_\text{view}$; and \emph{Random}.
% Extreme occlusion is held out from training to test generalization.
% Curves show the mean over $100$, $100$, and $50$ scenes for the low-, high-, and extreme-occlusion settings, respectively.
% DiVE achieves the best reconstruction across all settings, with its margin widening as occlusion increases.}
%     \label{fig:main_result}
%     \vspace{-12pt}
% \end{wrapfigure}

\subsection{Main Comparison}
\label{sec:main_comparison}
\vspace{-5pt}

% We compare DiVE (Ours) against three baselines.
% \textbf{CNABU}~\citep{Marques2025} is the manipulation-enhanced mapping planner of Marques et al., which selects view and push actions by maximizing volumetric information gain over a two-step lookahead, evaluated through a calibrated neural belief update that predicts each candidate's post-action belief.
% \textbf{View-Only} is an ablation of DiVE that always selects the viewing mode while retaining $\pi_\text{view}$, isolating the role of push-resolvable visibility.
% \textbf{Random} uniformly samples viewpoints and integrates observations with a classical occupancy update~\citep{Thrun2005}.
We compare DiVE (Ours) against three baselines.
\textbf{CNABU}~\citep{Marques2025} selects view and push actions by evaluating volumetric information gain from predicted post-action beliefs for all candidate actions. 
We train CNABU end-to-end on our dataset and evaluate it with the same action budget, candidate viewpoints, and candidate pushes as DiVE for a fair comparison.
\textbf{View-Only} is an ablation of DiVE that always selects the viewing mode while retaining $\pi_\text{view}$, isolating the role of push-resolvable visibility.
\textbf{Random} uniformly samples viewpoints and integrates observations with a classical occupancy update~\citep{Thrun2005}; it is included only as a lower-bound exploration baseline rather than as an action-selection ablation.

% Figure~\ref{fig:results}(a) reports occupancy IoU and semantic mIoU over the action budget.
% In low-occlusion scenes all informed methods perform similarly, since most unobserved regions are recoverable by changing viewpoints.
% As occlusion increases, View-Only and Random plateau early while the push-enabled methods keep improving, showing that some regions stay hidden from every viewpoint and can be exposed only through interaction.

Figure~\ref{fig:results}(a) reports occupancy IoU and semantic mIoU over the action budget.
In low-occlusion scenes, all informed methods perform similarly because most hidden regions can be recovered by viewpoint changes alone.
As occlusion increases, View-Only and Random plateau early, while push-enabled methods continue improving, confirming that some regions are not view-resolvable and require physical interaction.
% DiVE leads CNABU in occupancy throughout the budget under high and extreme occlusion, and overtakes it in semantic mIoU under extreme occlusion as actions accumulate.
% CNABU rolls out each candidate's post-action belief and scores its information gain, which makes action values sensitive to errors in the predicted belief when much of the scene is unobserved.
% DiVE instead predicts decomposed visibility maps, $\phi_\text{view}$ and $\phi_\text{push}$, and compares viewing and pushing by their expected visibility change---asking whether hidden regions are view- or push-resolvable rather than predicting how the full belief will evolve.
% Its advantage therefore widens with occlusion and persists under extreme occlusion, outside the training distribution, indicating that reasoning over visibility transfers to harder clutter than seen in training.
% DiVE outperforms CNABU most clearly under high and extreme occlusion.
% CNABU evaluates candidate actions by rolling out predicted post-action beliefs and scoring uncertainty gain, which becomes sensitive to belief prediction errors when large portions of the scene are unobserved.
% In contrast, DiVE selects actions from decomposed visibility maps, asking whether hidden regions are view-resolvable or push-resolvable rather than predicting the full future belief.
% This visibility-centered decision becomes more useful as occlusion increases, allowing DiVE to maintain its advantage even under extreme occlusion, which is held out from training.
DiVE outperforms CNABU most clearly under high and extreme occlusion.
CNABU computes volumetric information gain for all candidate actions from their predicted post-action beliefs, which becomes sensitive to belief prediction errors when large portions of the scene are unobserved.
In contrast, DiVE selects actions from decomposed visibility maps, asking whether hidden regions are view-resolvable or push-resolvable rather than predicting the full future belief.
This visibility-centered decision becomes more useful as occlusion increases, allowing DiVE to maintain its advantage even under extreme occlusion, an occlusion regime not seen during training.
In extreme occlusion, DiVE initially trails CNABU but surpasses it as actions accumulate.
% This reflects its hierarchical selection: rather than switching to pushing based on a single promising push candidate, DiVE compares mode-level expected gains and continues gathering view information while viewing remains more informative.
This behavior is consistent with its hierarchical selection: rather than switching to pushing based on a single promising push candidate, DiVE compares mode-level expected gains and can continue gathering view information while viewing remains informative.

DiVE is also substantially faster than CNABU.
As shown in Fig.~\ref{fig:results}(c), DiVE reduces per-step planning time by $19.8\times$ and total per-step time by $13.2\times$.
This speedup comes from replacing CNABU's 
exhaustive candidate-wise computation of 
post-action belief gains with a mode-specific policy that ranks candidates within the selected mode using a single forward pass.

\vspace{-8pt}

\subsection{Action-Selection Analysis}
\label{exp:action_selection}
\vspace{-5pt}
\paragraph{Selection quality.}
\begin{wrapfigure}{r}{0.32\linewidth}
    \vspace{-12pt}
    \centering
    \includegraphics[width=0.96\linewidth]
    {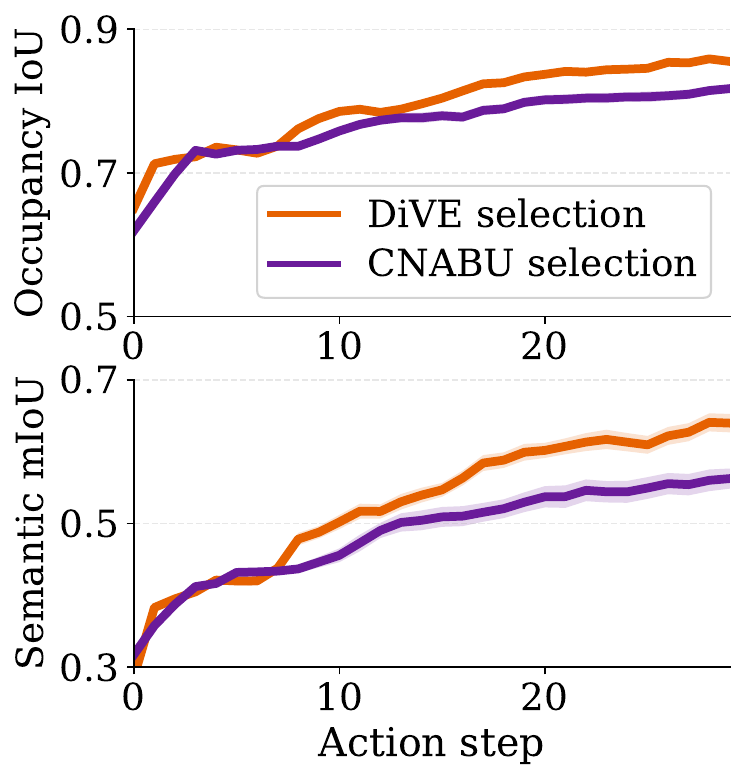}
    \vspace{-8pt}
    \caption{\textbf{Selection quality.}
    Reconstruction quality with the belief estimator fixed and only the
    action-selection strategy changed.}
    \label{fig:selection_quality}
    \vspace{-15pt}
\end{wrapfigure}

To isolate the effect of action selection, we fix the belief estimator $f_\text{scene}$ and compare decomposed-visibility selection with belief-gain selection (Fig.~\ref{fig:selection_quality}).
Since both variants use the same belief estimator, the performance gap reflects action-selection quality rather than reconstruction-module differences.
DiVE achieves higher reconstruction performance, showing that decomposed visibility leads to more informative action choices than the belief-gain criterion.

\vspace{-5pt}
\paragraph{Visibility decomposition.}
\label{exp:decomposition}
DiVE represents visibility using two separate maps, $\phi_\text{view}$ and $\phi_\text{push}$, rather than a single unified map.
The unified baseline combines estimated view-resolvable visibility change and push-resolvable gain in one channel; 
implementation details are provided in the Appendix.
For each candidate push with swept volume $\Omega_p$, 
we compute a map-based score $s(p)=\sum_{v\in\Omega_p}\phi(v)$ using either $\phi_\text{push}$ or the unified map, and correlate this score with the true push gain over 76,721 policy-training samples.
Scores from $\phi_{\text{push}}$ correlate more strongly with true push gain than scores from the unified map (0.436 vs. 0.204). The full distribution is shown in the Appendix.
This shows that decomposing visibility preserves the push-specific cue needed for candidate ranking, whereas mixing view- and push-resolvable signals dilutes this cue and weakens the subsequent view-push decision.
\vspace{-5pt}

\subsection{Robustness}
\label{sec:robustness}
\vspace{-5pt}

\paragraph{Action-space and planner sensitivity.}
% Figure~\ref{fig:analysis_summary}(b--c) varies one setting at a time from the default action space and two-view warm-up.
% Across action-space variants, $|\Delta\mathrm{IoU}|\leq1.75$ percentage points.
% Fixed schedules reduce high-occlusion IoU by 0.45--1.19 points, while longer warm-ups provide no improvement, supporting learned gain-based mode selection.
\begin{wraptable}{r}{0.62\linewidth}
\vspace{-12pt}
\centering
\captionsetup{width=\linewidth}

% ---------------------------------------------------------
% (a) Action-space sensitivity
% ---------------------------------------------------------
\begin{subtable}[t]{0.68\linewidth}
\centering

{\fontsize{6.2}{7.1}\selectfont
\setlength{\tabcolsep}{2.2pt}

\begin{tabular*}{\linewidth}
{@{\extracolsep{\fill}}llrr@{}}
\toprule
Parameter (default) & Variant & High & Ext. \\
\midrule

View candidates (98)
    & 49
    & $-1.01$
    & $-0.45$ \\

    & 24
    & $+1.06$
    & $+1.49$ \\

View span (full)
    & Narrow (49)
    & $-1.32$
    & $-0.68$ \\

Push candidates (20)
    & 10
    & $-0.97$
    & $-0.35$ \\

    & 40
    & $+0.17$
    & $+0.99$ \\

Push length (15~cm)
    & 12~cm
    & $-1.46$
    & $-0.76$ \\

    & 18~cm
    & $-1.24$
    & $+0.14$ \\

Direction ($\pm30^\circ$)
    & $\pm20^\circ$
    & $-1.75$
    & $+0.50$ \\

\bottomrule
\end{tabular*}
}

\vspace{-3pt}
\caption{}
\label{tab:action_space_sensitivity}
\end{subtable}
\hfill
%
% ---------------------------------------------------------
% (b) Planner sensitivity
% ---------------------------------------------------------
\begin{subtable}[t]{0.29\linewidth}
\centering

{\fontsize{6.2}{7.1}\selectfont
\setlength{\tabcolsep}{2.0pt}

\begin{tabular*}{\linewidth}
{@{\extracolsep{\fill}}lr@{}}
\toprule
Alternative (High) & $\Delta$IoU \\
\midrule

\multicolumn{2}{@{}l}{\itshape Fixed push interval} \\

Push every 3 & $-0.86$ \\
Push every 5 & $-1.19$ \\
Push every 7 & $-0.45$ \\

\addlinespace[1pt]
\multicolumn{2}{@{}l}{\itshape Post-push warm-up} \\

Warm-up 3 & $-0.68$ \\
Warm-up 4 & $-1.76$ \\
Warm-up 5 & $-1.26$ \\

\bottomrule
\end{tabular*}
}

\vspace{-3pt}
\caption{}
\label{tab:planner_sensitivity}
\end{subtable}

\vspace{-5pt}

\caption{\textbf{Robustness analyses.}
\textbf{(a)} Final-IoU change (percentage points) when varying one
action-space parameter from the parenthesized default.
\textbf{(b)} Final-IoU change under fixed push schedules or longer
warm-ups; the default uses learned mode selection and a two-view
warm-up.}

\label{tab:robustness_analysis}
\vspace{-12pt}
\end{wraptable}
We vary one parameter at a time from the default action space of 98 viewpoints, 20 push candidates, a 15~cm push length, and a $\pm30^\circ$ direction range.
Across all variants, the final occupancy IoU changes by at most 1.75 percentage points on high- and extreme-occlusion scenes (Table~\ref{tab:robustness_analysis}(a)), indicating that performance is not tied to a particular action-space parameterization.
% We also replace the learned mode selector with fixed schedules that push every 3, 5, or 7 actions; all reduce high-occlusion IoU by 0.45--1.19 points (Fig.~\ref{fig:analysis_summary}(c)).
% Extending the post-push warm-up from two views to three, four, or five also decreases IoU by 0.68--1.76 points, supporting learned mode selection with the two-view warm-up.
We also examine the sensitivity of the planner design. 
To avoid mode selection from an under-observed visibility state, we use a two-view warm-up at the beginning of each episode and after every push, and activate learned mode selection only thereafter. 
We replace the learned mode selector with fixed schedules that push every 3, 5, or 7 actions; all reduce high-occlusion IoU by 0.45–1.19 points (Table~\ref{tab:robustness_analysis}(b)). 
Extending the post-push warm-up from two views to three, four, or five also decreases IoU by 0.68–1.76 points. 
Together, these results support adaptive mode selection with the two-view warm-up used in our main experiments.
\vspace{-5pt}
% \paragraph{Push execution reliability.}
% We consider a push successful when it contacts the intended object without tipping it or moving it outside the workspace.
% Under this criterion, 180 of 181 simulation pushes succeed (99.4\%), as do all pushes performed across the 10 real-world episodes.
% Because failures are rare under our constrained push primitive, the current planner ranks actions by visibility gain without explicitly estimating execution probability.
% For more challenging settings, the gain could instead be weighted by predicted success probability, cost, or safety risk.

\subsection{Generalization}
\label{sec:ood}
\vspace{-5pt}
We evaluate whether DiVE remains effective under two object distribution shifts (Fig.~\ref{fig:results}(b)).
For \emph{shape}, we rescale every object in the low-occlusion test set by a random factor in $[0.8, 1.2]$ and evaluate zero-shot.
For \emph{category}, we introduce two unseen object categories and lightly finetune only the belief estimator on 30 low-occlusion scenes, keeping the visibility and action-selection modules frozen.
DiVE maintains similar reconstruction quality across both shifts.
These results suggest that DiVE generalizes zero-shot to object shape variation, while unseen categories can be handled with lightweight adaptation of the reconstruction module rather than retraining the full framework.

\vspace{-5pt}
\subsection{Real-World Experiments}
\label{sec:real_world}
\vspace{-5pt}
We deploy DiVE zero-shot on the Piper AgileX setup described in Sec.~\ref{sec:setup}. 
We construct 10 real cluttered shelf scenes using YCB-category objects whose shapes and materials differ from simulation, and compare against View-Only over a 20-action budget. 
We report object-level discovery metrics: correctly found (CF), misclassified but found (MBF), missed, and initially occluded object discovery (Occ. Disc.).
The real-world setup and metrics are detailed in the Appendix.
%The real-world setup and metrics are detailed in Appendix~\ref{app:real_world}.

\begin{wraptable}{r}{0.50\linewidth}
\vspace{-10pt}
\centering

\begin{minipage}{\linewidth}
\centering
\captionsetup{width=\linewidth}

\setlength{\tabcolsep}{1.8pt}
\renewcommand{\arraystretch}{0.92}

{\fontsize{7.0}{8.0}\selectfont
\begin{tabular*}{\linewidth}
{@{\extracolsep{\fill}}lcccc@{}}
    \toprule
    Method
    & CF$\uparrow$
    & MBF
    & Missed$\downarrow$
    & Occ. Disc.$\uparrow$ \\
    \midrule

    View-only
    & $54.3{\pm}10.2$
    & $0.0{\pm}0.0$
    & $45.7{\pm}10.2$
    & $54.2{\pm}35.4$ \\

    \textbf{Ours}
    & $\mathbf{65.5{\pm}14.2}$
    & $5.4{\pm}6.9$
    & $\mathbf{29.1{\pm}15.4}$
    & $\mathbf{79.2{\pm}23.3}$ \\

    \midrule
    $\Delta$
    & $+11.2$
    & $+5.4$
    & $-16.6$
    & $+25.0$ \\
    \bottomrule
\end{tabular*}
}

\vspace{-4pt}
\caption{Zero-shot real-world results over 10 scenes.
All values are percentages.}
\label{tab:realworld_eval}

\end{minipage}
\vspace{-10pt}
\end{wraptable}

% Table~\ref{tab:realworld_eval} shows that DiVE improves reconstruction under this real-world sim-to-real gap. 
% It increases correctly found objects from $54.3\%$ to $65.5\%$ and reduces missed objects from $45.7\%$ to $29.1\%$. 
% The gain is largest for initially occluded objects, where DiVE discovers $79.2\%$ compared with $54.2\%$ for View-Only, suggesting that its learned action selection transfers to real scenes by using interaction to expose regions that viewpoint changes alone cannot reveal. 
% The small MBF increase indicates that a few objects newly exposed by pushing are geometrically recovered but semantically mislabeled; View-Only often leaves these regions empty, counting them as missed instead. 
% Overall, the results support the zero-shot transfer of visibility-based view-and-push exploration.

% \subsection{Real-World Experiments}
% \label{sec:real_world}
% \vspace{-5pt}
% We deploy DiVE zero-shot on the Piper AgileX setup described in Sec.~\ref{sec:setup}. 
% We construct 10 real cluttered shelf scenes using YCB-category objects whose shapes and materials differ from simulation, and compare against View-Only over a 20-action budget. 
% We report object-level discovery metrics: correctly found (CF), misclassified but found (MBF), missed, and initially occluded object discovery (Occ. Disc.).
% The real-world setup and metrics are detailed in Appendix~\ref{app:real_world}.

Table~\ref{tab:realworld_eval} shows that DiVE improves reconstruction after zero-shot transfer to real shelves. 
It increases correctly found objects from $54.3\%$ to $65.5\%$ and reduces missed objects from $45.7\%$ to $29.1\%$. 
The largest gain is in initially occluded objects, where DiVE discovers $79.2\%$ compared with $54.2\%$ for View-Only, suggesting that its learned action selection transfers to real scenes by choosing interactions that expose regions unreachable by viewpoint changes alone. 
The nonzero MBF rate reflects cases where pushing exposes objects that are geometrically recovered but semantically mislabeled; View-Only instead leaves many of these regions empty, so they are counted as missed.
Overall, these results demonstrate the real-world transferability of visibility-based view-and-push exploration.

\section{Conclusion}
\vspace{-8pt}
We presented DiVE, a framework for active scene reconstruction in cluttered environments that explicitly decomposes scene visibility into two complementary components: view-resolvable visibility \(\phi_{\mathrm{view}}\) and push-resolvable visibility \(\phi_{\mathrm{push}}\).
We adapt an established hierarchical view-and-push planner to this decomposed representation by mapping the modality-specific signals onto a common expected-visibility-change scale for mode selection. 
We further distill within-mode candidate gains into mode-specific policies for lightweight action ranking, avoiding candidate-wise post-action belief prediction and a hand-tuned cross-mode threshold. 
Experiments in both simulation and real-world cluttered shelf scenes demonstrate that, compared with prior methods based on unified belief representations, lookahead planning, or heuristic mode switching, DiVE selects more informative actions, reduces planning cost, and achieves the largest reconstruction gains under severe occlusion.
\vspace{-5pt}
\paragraph{Limitations and future work.}
Evaluation is limited to one cluttered-shelf domain, discrete action candidates, and no explicit execution-cost or risk model.
The grasp-resolvable head in Sec.~\ref{sec:robustness}, which predicts visibility changes induced by grasping an object, demonstrates at the representation level that DiVE can be extended to additional interaction primitives.
Future work should study risk-aware gain, diverse layouts and interaction primitives, open-vocabulary semantics, and downstream manipulation.

%===============================================================================

\clearpage
\acknowledgments{The authors thank the anonymous reviewers and the Area Chair for their constructive feedback, which helped improve the clarity of the paper and motivated additional analyses of action-space sensitivity, push reliability, and extensibility to additional manipulation primitives. We also thank our colleagues for their helpful discussions and feedback.

This work was supported by a National Research Foundation of Korea (NRF) grant funded by the Korean government (MSIT) (No. 2022M3C1A3099336). This work was also supported by the Institute of Information \& Communications Technology Planning \& Evaluation (IITP) under the Artificial Intelligence Semiconductor Support Program to Nurture the Best Talents (No. IITP-2023-RS-2023-00256081), funded by the Korean government (MSIT).
}

%===============================================================================

\bibliography{references}  % .bib
\clearpage 
\appendix

\appendix

\section{Methodological Details}
\label{app:methodology}

\subsection{Algorithm}
\label{app:algorithm}
\begin{algorithm}[H]
\caption{Active Scene Reconstruction with Decomposed Visibility}
\label{alg:overview}
\begin{algorithmic}[1]
\State Initialize $\phi_{\text{view},-1}, \phi_{\text{push},-1} \gets \mathbf{0}$
\State Initialize $\phi_{\text{scene},-1} \gets \mathbf{0.5}$ and $\Omega_0 \gets \mathbf{0}$
\State Acquire initial observation $o_0$ from viewpoint $k_0$
\State Set $\mathcal{K}_0 \gets \{k_0\}$ \Comment{viewpoints visited since the most recent push}

\For{$t = 0, 1, \dots, T-1$}
    \State \textcolor{gray}{\textit{// (i) Visibility and belief update}}
    \State $(\phi_{\text{view},t}, \phi_{\text{push},t}) \gets 
    f_{\text{DiVE}}(\phi_{\text{view},t-1}, \phi_{\text{push},t-1}, o_t)$
    \State $\phi_{\text{scene},t} \gets 
    f_{\text{scene}}(\phi_{\text{scene},t-1}, o_t, \Omega_t)$

    \State \textcolor{gray}{\textit{// (ii) Mode selection}}
    \State $\mathcal{G}_{\text{view}} \gets 
    \sum_{v \in \mathcal{V}}
    \dfrac{
    2\,\phi_{\text{view},t}(v)\bigl(1-\phi_{\text{view},t}(v)\bigr)
    }{
    |\mathcal{K}_t| + 1
    }$
    \State $\mathcal{G}_{\text{push}} \gets 
    \sum_{v \in \mathcal{V}} \phi_{\text{push},t}(v)$
    \If{$|\mathcal{K}_t| < 2$}
        \State $\mathcal{M}_t \gets \text{view}$ \Comment{two-view warm-up}
    \Else
        \State $\mathcal{M}_t \gets 
        \arg\max_{m \in \{\text{view}, \text{push}\}} \mathcal{G}_m$
    \EndIf

    \State \textcolor{gray}{\textit{// (iii) Action selection}}
    \If{$\mathcal{M}_t = \text{view}$}
        \State $a_t \gets \pi_{\text{view}}(\phi_{\text{view},t}, \mathcal{K}_t)$
        \State Execute viewpoint action $a_t$ and acquire $o_{t+1}$ from viewpoint $k_{t+1}=a_t$
        \State $\Omega_{t+1} \gets \mathbf{0}$
        \State $\mathcal{K}_{t+1} \gets \mathcal{K}_t \cup \{k_{t+1}\}$
    \Else
        \State $a_t \gets \pi_{\text{push}}(\phi_{\text{push},t}, \mathcal{P}_t)$
        \State Execute push action $a_t$ and acquire $o_{t+1}$ from viewpoint $k_{t+1}$
        \State $\Omega_{t+1} \gets \Omega(a_t)$
        \State $\mathcal{K}_{t+1} \gets \{k_{t+1}\}$ \Comment{reset after the scene geometry changes}
    \EndIf
\EndFor

\State \Return $\phi_{\text{scene},T}$
\end{algorithmic}
\end{algorithm}

\subsection{Derivation and Empirical Validation of the View-Mode Gain Proxy}
\label{app:view_change_derivation}

We derive the analytic view-mode gain proxy used in Eq.~\ref{eq:view_change} and empirically validate whether it matches the data-derived visibility change from actual candidate viewpoints. This validation is important because the planner compares viewing and pushing on a common expected-visibility-change scale. The view proxy therefore does not need to predict the exact effect of every candidate viewpoint, but it should provide a stable estimate of whether additional viewing is still expected to change the visibility state.

\paragraph{Setup.}
Fix a voxel $v \in \mathcal{V}$. Let $\mathcal{K}_t$ denote the set of viewpoints visited since the most recent push, and let $n \coloneqq |\mathcal{K}_t|$. The reset after a push is necessary because pushing changes the scene geometry; visibility statistics accumulated before the push no longer describe the same physical arrangement.

Let $m_k(v) \in \{0,1\}$ denote the binary visibility of voxel $v$ from viewpoint $k$. The view-resolvable visibility state is defined as the running average of the binary visibility maps over the accumulated viewpoints:
\begin{equation}
\phi_{\text{view},t}(v)
=
\frac{1}{n}
\sum_{k \in \mathcal{K}_t} m_k(v).
\label{eq:app_phi_running_avg}
\end{equation}
Let $k'$ denote one additional viewpoint to be observed. We define
\begin{equation}
X \coloneqq m_{k'}(v) \in \{0,1\}
\end{equation}
as the Bernoulli random variable indicating whether voxel $v$ is visible from this additional viewpoint.
After incorporating this additional viewpoint, the updated running average becomes
\begin{equation}
\phi_{\text{view},t+1}(v)
=
\frac{1}{n+1}
\left(
\sum_{k \in \mathcal{K}_t} m_k(v) + X
\right).
\label{eq:app_phi_running_avg_next}
\end{equation}

\paragraph{Derivation.}
Let
\begin{equation}
S \coloneqq \sum_{k \in \mathcal{K}_t} m_k(v).
\end{equation}
Subtracting Eq.~\ref{eq:app_phi_running_avg} from Eq.~\ref{eq:app_phi_running_avg_next}, we obtain
\begin{align}
\phi_{\text{view},t+1}(v) - \phi_{\text{view},t}(v)
&=
\frac{S+X}{n+1} - \frac{S}{n} \\
&=
\frac{n(S+X) - (n+1)S}{n(n+1)} \\
&=
\frac{nX-S}{n(n+1)}.
\end{align}
Since $S = n\,\phi_{\text{view},t}(v)$,
\begin{align}
\phi_{\text{view},t+1}(v) - \phi_{\text{view},t}(v)
&=
\frac{nX - n\,\phi_{\text{view},t}(v)}{n(n+1)} \\
&=
\frac{X-\phi_{\text{view},t}(v)}{n+1}.
\label{eq:app_phi_diff}
\end{align}
Taking the absolute value gives
\begin{equation}
\left|
\phi_{\text{view},t+1}(v) - \phi_{\text{view},t}(v)
\right|
=
\frac{
\left|X-\phi_{\text{view},t}(v)\right|
}{
n+1
}.
\label{eq:app_phi_abs_diff}
\end{equation}

We next use $\phi_{\text{view},t}(v)$ as a plug-in estimate of the Bernoulli mean for the visibility of voxel $v$ from an additional viewpoint:
\begin{equation}
\Pr(X=1) = \phi_{\text{view},t}(v), \qquad
\Pr(X=0) = 1-\phi_{\text{view},t}(v).
\end{equation}
Writing $p \coloneqq \phi_{\text{view},t}(v)$ for brevity,
\begin{align}
\mathbb{E}\left[\left|X-p\right|\right]
&=
\Pr(X=1)|1-p| + \Pr(X=0)|0-p| \\
&=
p(1-p) + (1-p)p \\
&=
2p(1-p).
\end{align}
Substituting this into Eq.~\ref{eq:app_phi_abs_diff}, we obtain
\begin{equation}
\mathbb{E}_{k'}\left[
\left|
\phi_{\text{view},t+1}(v) - \phi_{\text{view},t}(v)
\right|
\right]
=
\frac{
2\,\phi_{\text{view},t}(v)
\left(
1-\phi_{\text{view},t}(v)
\right)
}{
n+1
}
=
\frac{
2\,\phi_{\text{view},t}(v)
\left(
1-\phi_{\text{view},t}(v)
\right)
}{
|\mathcal{K}_t|+1
}.
\label{eq:app_view_proxy}
\end{equation}
We denote this analytic plug-in estimate as
\begin{equation}
A_n(v)
\coloneqq
\frac{
2\,\phi_{\text{view},t}(v)
\left(
1-\phi_{\text{view},t}(v)
\right)
}{
n+1
}.
\label{eq:app_an_view_proxy}
\end{equation}

\paragraph{Interpretation.}
The term $2p(1-p)$ arises from the expected absolute deviation of a Bernoulli variable from its mean, and is equal to twice the variance of a Bernoulli variable with mean $p$. It is maximized at $p=0.5$ and vanishes as $p$ approaches either $0$ or $1$. Therefore, the proxy assigns the largest expected update to voxels whose visibility is inconsistent across viewpoints, and assigns small values to voxels that are consistently visible or consistently invisible. The denominator $n+1$ captures the diminishing influence of one additional viewpoint as more viewpoints are accumulated. Thus, $A_n(v)$ estimates how much one more view is expected to change the current visibility state, not whether the voxel is occupied or semantically correct.

\paragraph{Empirical validation protocol.}
The derivation above uses a plug-in Bernoulli model. To test whether this approximation is useful on our data, we compare the analytic proxy with the empirical visibility-state change computed from actual candidate-view visibility maps.

For each validation instance, we construct a partial viewpoint set $\mathcal{K}_t$ of size $n$ and compute $\phi_{\text{view},t}$ using Eq.~\ref{eq:app_phi_running_avg}. We then evaluate the actual effect of adding one more candidate viewpoint using the ground-truth binary visibility maps:
\begin{equation}
\Delta^{\mathrm{emp}}_n(v)
=
\frac{1}{|\mathcal{C}|}
\sum_{k' \in \mathcal{C}}
\left|
\frac{
n\,\phi_{\text{view},t}(v) + m_{k'}(v)
}{
n+1
}
-
\phi_{\text{view},t}(v)
\right|.
\label{eq:app_emp_view_change}
\end{equation}
This quantity is the data-derived expected absolute update of the running visibility state after one additional viewpoint.

\begin{figure}[t]
\centering
\includegraphics[width=0.90\linewidth]{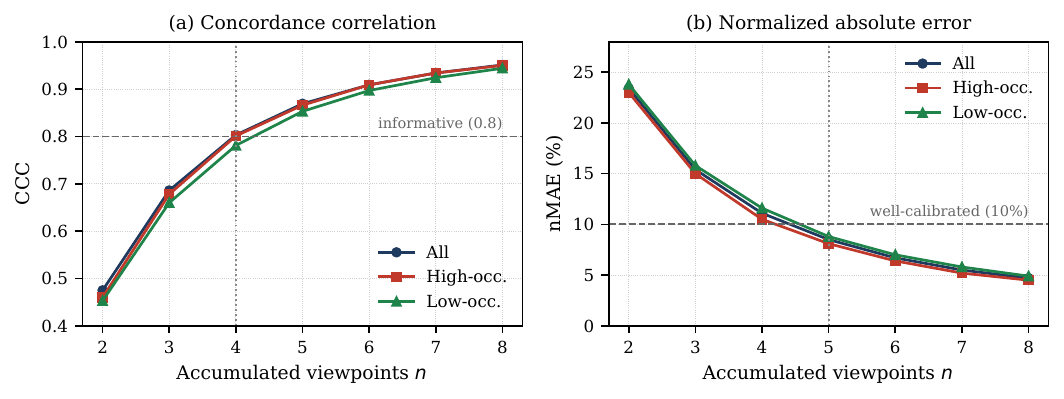}
\caption{
Empirical validation of the analytic view-mode gain proxy. We compare the analytic proxy $\widehat{G}^{(n)}_{\text{view}}$ with the data-derived empirical visibility-state change $G^{(n),\mathrm{emp}}_{\text{view}}$ as the number of accumulated viewpoints $n$ increases. The proxy is noisy for the first few observations, but becomes informative around $n=4$ and well calibrated from $n=5$ onward. The same trend appears in both low- and high-occlusion scenes, indicating that the approximation is not specific to one occlusion regime.
}
\label{fig:view_proxy_validation}
\end{figure}

Since mode selection uses the voxel-summed gain rather than individual voxel values, we compare the corresponding map-level quantities:
\begin{equation}
\widehat{G}^{(n)}_{\text{view}}
=
\sum_{v\in\mathcal{V}} A_n(v),
\qquad
G^{(n),\mathrm{emp}}_{\text{view}}
=
\sum_{v\in\mathcal{V}} \Delta^{\mathrm{emp}}_n(v).
\label{eq:app_view_gain_emp_compare}
\end{equation}
We report Pearson correlation, concordance correlation coefficient (CCC), $R^2$ with respect to the identity line $y=x$, normalized mean absolute error (nMAE), and normalized bias between $\widehat{G}^{(n)}_{\text{view}}$ and $G^{(n),\mathrm{emp}}_{\text{view}}$. We evaluate this agreement for $n=1,\dots,8$ on all validation samples, and separately on low- and high-occlusion scenes. Figure~\ref{fig:view_proxy_validation} visualizes the agreement between the analytic proxy and the empirical view-induced visibility change as the number of accumulated viewpoints increases. Table~\ref{tab:view_proxy_validation} reports representative numerical values from this validation.

\begin{table}[t]
\centering
\small
\setlength{\tabcolsep}{4pt}        % horizontal padding between columns
\renewcommand{\arraystretch}{1.08} % vertical spacing between rows
\caption{
Empirical validation of the view-mode gain proxy. The proxy becomes informative after a few accumulated viewpoints: at $n=4$, CCC is around $0.80$ with nMAE near $11\%$, and from $n=5$ onward nMAE falls below $10\%$ across both low- and high-occlusion scenes. Full trends over $n=1,\dots,8$ are shown in Fig.~\ref{fig:view_proxy_validation}.
}
\label{tab:view_proxy_validation}
\begin{tabular*}{\linewidth}{@{\extracolsep{\fill}}lcccccc@{}}
\toprule
Split & $n$ & $N$ & Pearson & CCC & $R^2_{y=x}$ & nMAE \\
\midrule
All & 3 & 12.8k & 0.877 & 0.686 & 0.212 & 15.4\% \\
All & 4 & 12.7k & 0.920 & 0.803 & 0.558 & 11.1\% \\
All & 5 & 12.8k & 0.941 & 0.869 & 0.718 & 8.5\% \\
All & 8 & 12.7k & 0.970 & 0.951 & 0.901 & 4.7\% \\
\midrule
High-occ. & 4 & 6.4k & 0.915 & 0.801 & 0.548 & 10.5\% \\
High-occ. & 5 & 6.5k & 0.936 & 0.866 & 0.710 & 8.1\% \\
Low-occ. & 4 & 6.4k & 0.915 & 0.781 & 0.489 & 11.6\% \\
Low-occ. & 5 & 6.4k & 0.937 & 0.853 & 0.676 & 8.8\% \\
\bottomrule
\end{tabular*}
\end{table}

\paragraph{Discussion.}
The plug-in proxy is less reliable when only one or two viewpoints have been accumulated, which is expected because $\phi_{\text{view},t}$ is then estimated from very sparse visibility observations. However, its agreement with the empirical visibility-state change improves rapidly after a few views. Over all validation samples, CCC increases from $0.686$ at $n=3$ to $0.803$ at $n=4$, while nMAE decreases from $15.4\%$ to $11.1\%$. From $n=5$ onward, nMAE falls below $10\%$, and by $n=8$ the proxy reaches CCC $=0.951$ and nMAE $=4.7\%$.

The same trend holds under both occlusion regimes. At $n=4$, high- and low-occlusion scenes obtain CCC values of $0.801$ and $0.781$, respectively. At $n=5$, nMAE is below $9\%$ in both splits. This indicates that the proxy is not specific to a single visibility regime, but provides a consistent estimate of expected view-induced visibility change across different clutter densities.

These results support using $A_n$ as a mode-selection proxy rather than as an exact per-voxel predictor. The planner compares the voxel-summed view gain $\widehat{G}^{(n)}_{\text{view}}$ with the push gain, so individual voxel-level errors are averaged over the scene. Therefore, the proxy only needs to estimate whether additional viewing is still expected to meaningfully change the visibility state at the mode level. Empirically, this condition is reasonably met around $n=4$ and becomes more reliable from $n=5$ onward. This matches the intended use of the proxy in the hierarchical planner: viewing is preferred while the accumulated visibility state can still be meaningfully updated, whereas pushing becomes favorable when the expected gain from additional views is limited relative to the predicted push-induced visibility change.

\section{Implementation Details}
\label{app:implementation}

\subsection{Simulation Data Generation}
\label{app:data_generation}

We generate cluttered shelf scenes in Isaac Sim using a single shelf workspace and objects drawn from the YCB object set. The in-distribution object categories used for training and standard evaluation are listed in Table~\ref{tab:ycb_objects}. The model is trained on low- and high-occlusion scenes, while extreme-occlusion scenes are held out for evaluation as described in Sec.~\ref{app:test_scene_generation}.

\begin{table}[h]
\centering
\small
\caption{YCB objects used in the in-distribution cluttered shelf scenes.}
\label{tab:ycb_objects}
\begin{tabular}{cl}
\toprule
Index & YCB object \\
\midrule
0 & 003\_cracker\_box \\
1 & 004\_sugar\_box \\
2 & 006\_mustard\_bottle \\
3 & 021\_bleach\_cleanser \\
4 & 002\_master\_chef\_can \\
5 & 005\_tomato\_soup\_can \\
6 & 009\_gelatin\_box \\
7 & 010\_potted\_meat\_can \\
8 & 008\_pudding\_box \\
9 & 007\_tuna\_fish\_can \\
10 & 061\_foam\_brick \\
\bottomrule
\end{tabular}
\end{table}

\paragraph{Voxel representation.}
Each scene is discretized into $\mathcal{V} \subset \mathbb{R}^{120{\times}60{\times}80}$, covering a $60{\times}30{\times}40$~cm workspace in front of the shelf. Each voxel is therefore approximately $5$~mm on a side. We use the same voxel grid for visibility estimation, action selection, belief-map updating, and evaluation.

\paragraph{Candidate viewpoints.}
The candidate viewpoint set $\mathcal{C}$ contains 98 camera poses arranged on a $7{\times}14$ planar grid in front of the shelf. Each camera is oriented toward the center of the shelf so that the candidate set covers different viewing angles over the same workspace. We index viewpoints as
\begin{equation}
\text{cam\_idx} = \text{row} \times 14 + \text{col}.
\end{equation}
All methods use the same calibrated RGB-D camera model and the same candidate viewpoint set. Thus, performance differences arise from action selection and belief updating rather than from changes in sensing geometry.

\paragraph{Candidate pushes and swept maps.}
At each decision step, we construct a candidate push set $\mathcal{P}_t$ with 20 actions. Candidate contacts are sampled from frontier regions near the boundary between observed occupied space and unobserved space. For each contact, we sample a pushing direction within $\pm 30^\circ$ of the shelf depth axis and apply a fixed push length of $15$~cm. Each push candidate $p \in \mathcal{P}_t$ is represented by its swept map $\Omega_p$, which marks the voxels physically affected by the push. The swept map is used by the push policy and is passed to the belief map estimator after executing an interaction.

\paragraph{Episode data.}
During data generation, we render all 98 candidate viewpoints for each scene state. This allows us to compute dense supervision for view-resolvable visibility, push-resolvable visibility, and candidate-level policy targets without repeatedly simulating online planning rollouts. Push sequences contain multiple interaction states, and all samples derived from the same initial scene inherit the same train/validation split to prevent leakage.

\paragraph{Push supervision and execution handling.}
For every pre-push scene, we store the complete physical configuration and render all 98 candidate viewpoints to obtain the pre-push marginal visibility.
Before evaluating each push candidate, we restore this identical configuration, execute the candidate, and re-render all 98 viewpoints to compute its post-push marginal visibility.
Thus, candidate targets differ only in the effect of the evaluated push rather than in the scene state from which it starts.
Invalid candidates are masked during training and inference. After all candidates have been evaluated, one valid post-push configuration is sampled as the next state when generating multi-push sequences.

We define an executed push as successful when it contacts the intended object without tipping it or dropping it from the workspace; failures are detected from the resulting object configuration.
Under this criterion, 180 of 181 executed simulation pushes were successful (99.4\%), and all pushes executed in the 10 real-world episodes were successful.
Because failures were rare under the constrained primitive, the current selector does not explicitly predict success probability; a risk-aware extension could weight predicted visibility gain by execution success probability, cost, or safety.

\paragraph{Feasibility beyond pushing.}
To test whether the representation can accommodate another interaction primitive, we additionally train a grasp-resolvable visibility head using the marginal visibility change produced by removing a grasped object as supervision.
The learned head produces spatially localized grasp-resolvable responses, showing that the modality-conditioned representation can be instantiated for a third primitive.
This experiment is a representation-level proof of concept only: it does not evaluate closed-loop grasp selection, joint view--push--grasp episodes, or improvement on a downstream manipulation task.

\subsection{Network Architecture}
\label{app:architectures}

The implementation follows the decomposition introduced in the main paper: DiVE predicts two visibility maps, the mode-specific policies rank candidate actions, and the belief map estimator maintains the global reconstruction. We describe the modules using the paper notation rather than implementation-specific code names.

\paragraph{DiVE.}
DiVE, denoted by $f_{\text{DiVE}}$, is implemented as a shared volumetric backbone with two output heads. The input consists of the previous view-resolvable visibility state, the previous push-resolvable visibility state, and the current voxel observation:
\begin{equation}
[\phi_{\text{view},t-1},\phi_{\text{push},t-1},o_t^{\text{vox}}].
\end{equation}
The network outputs two voxel-wise logits:
\begin{equation}
(\ell_{\text{view},t},\ell_{\text{push},t})
=
f_{\text{DiVE}}(\phi_{\text{view},t-1},\phi_{\text{push},t-1},o_t^{\text{vox}}),
\end{equation}
which are converted to visibility maps by sigmoid activations:
\begin{equation}
\phi_{\text{view},t}=\sigma(\ell_{\text{view},t}),\qquad
\phi_{\text{push},t}=\sigma(\ell_{\text{push},t}).
\end{equation}
The view head is supervised with accumulated visibility frequency over visited viewpoints, whereas the push head is supervised with marginal visibility change induced by interaction. The swept map is not an input to DiVE; it is used by the push policy and by the belief map estimator after a push is executed.

\paragraph{View policy.}
The view policy $\pi_{\text{view}}$ ranks the 98 candidate viewpoints. It takes the current view-resolvable visibility map together with the viewpoint history $\mathcal{K}_t$, encoded as a seen-view mask. The policy outputs one logit per candidate viewpoint:
\begin{equation}
\pi_{\text{view}}(\phi_{\text{view},t},\mathcal{K}_t)
\rightarrow
\{\hat{\ell}_{\text{view}}(k)\}_{k\in\mathcal{C}}.
\end{equation}
At inference time, the next viewpoint is selected from the candidate set using the predicted scores.

\paragraph{Push policy.}
The push policy $\pi_{\text{push}}$ ranks the 20 candidate pushes. Each candidate is represented by its swept map $\Omega_p$, and invalid push candidates are masked during training and inference. The implementation scores each push using its overlap with the push-resolvable visibility map together with a learned residual term. Candidate-level self-attention is used so that candidate pushes can be compared within the same decision state. The policy outputs one logit per push:
\begin{equation}
\pi_{\text{push}}(\phi_{\text{push},t},\mathcal{P}_t)
\rightarrow
\{\hat{\ell}_{\text{push}}(p)\}_{p\in\mathcal{P}_t}.
\end{equation}
The selected push is the valid candidate with the highest predicted score.

\paragraph{Belief map estimator.}
The belief map estimator $f_{\text{scene}}$ maintains the global geometric and semantic reconstruction. Occupancy is represented by Beta parameters $(\alpha_t,\beta_t)$, and semantics are represented by voxel-wise semantic logits. The implementation uses $n_{\text{sem}}=15$ semantic channels: class $0$ denotes free space, 11 object classes are active in the in-distribution data, and three additional slots are reserved for unseen-class finetuning.

The input to the belief map estimator contains the previous occupancy belief, the current per-view occupancy observation, the swept map, the current semantic observation, and the previous semantic belief:
\begin{equation}
\left[
\alpha_{t-1}/50,\,
\beta_{t-1}/50,\,
o_t^{\text{occ}},\,
\Omega_t,\,
o_t^{\text{sem}},\,
\phi^{\text{sem}}_{\text{scene},t-1}
\right].
\end{equation}
The estimator outputs updated occupancy parameters and semantic logits:
\begin{equation}
(\alpha_t,\beta_t,\hat{s}_t)
=
f_{\text{scene}}(\phi_{\text{scene},t-1},o_t,\Omega_t).
\end{equation}
The semantic belief passed to the next step is computed from the detached softmax of $\hat{s}_t$. This stabilizes sequential belief updates and prevents gradients from propagating through long rollout histories.

\begin{table}[t]
\centering
\small
\setlength{\tabcolsep}{3pt}
\renewcommand{\arraystretch}{1.08}
\caption{Implementation-level module specification.}
\label{tab:module_io}
\begin{tabular}{@{}p{0.12\linewidth}p{0.25\linewidth}p{0.31\linewidth}p{0.23\linewidth}@{}}
\toprule
Module & Role & Input & Output \\
\midrule
$f_{\text{DiVE}}$ & Decomposed visibility estimation & Previous $\phi_{\text{view}}$, previous $\phi_{\text{push}}$, current voxel observation & Voxel-wise logits for $\phi_{\text{view}}$ and $\phi_{\text{push}}$ \\
$\pi_{\text{view}}$ & Viewpoint ranking & $\phi_{\text{view}}$ and seen-view mask & 98 candidate-view logits \\
$\pi_{\text{push}}$ & Push ranking & $\phi_{\text{push}}$, 20 swept maps, valid-candidate mask & 20 candidate-push logits \\
$f_{\text{scene}}$ & Belief map estimation & Beta occupancy belief, observation, swept map, semantic belief & Updated Beta occupancy parameters and 15-way semantic logits \\
\bottomrule
\end{tabular}
\end{table}

\subsection{Training Targets and Losses}
\label{app:losses}

\paragraph{DiVE.}
The view head is supervised by the accumulated visibility-frequency target in Eq.~\ref{eq:gt_view}. For a sequence of visited viewpoints, the target is the running average of per-view binary visibility maps. The push head is supervised by the push-resolvable visibility target in Eq.~\ref{eq:gt_push}, computed from the marginal visibility change induced by candidate pushes. We train DiVE with soft binary cross entropy:
\begin{equation}
\mathcal{L}_{\text{DiVE}}
=
\mathrm{BCEWithLogits}(\ell_{\text{view}},\phi^{*}_{\text{view}})
+
\lambda_{\text{push}}\,
\mathrm{BCEWithLogits}(\ell_{\text{push}},\phi^{*}_{\text{push}}),
\end{equation}
where $\lambda_{\text{push}}=1.0$.

\paragraph{View policy.}
The view policy is trained by distilling candidate-level visibility gains. For each training state, candidate gains $g_{\text{view}}(k)$ are computed by measuring how much the view-resolvable visibility state changes after adding candidate viewpoint $k$. Since the absolute gain scale varies across scenes, we normalize gains within each candidate set:

\begin{equation}
\tilde{g}_{\text{view}}(k)
=
\frac{
g_{\text{view}}(k)-\mu_{\mathcal{C}}(g_{\text{view}})
}{
\sigma_{\mathcal{C}}(g_{\text{view}})+\varepsilon
}.
\end{equation}
The normalized gains are converted into soft targets using a sigmoid with temperature $T$:
\begin{equation}
\tau_{\text{view}}(k)
=
\sigma\!\left(\frac{\tilde{g}_{\text{view}}(k)}{T}\right).
\end{equation}
% The policy is trained with binary cross entropy over the candidate-view logits:
The view policy outputs one logit $\hat{\ell}_{\text{view}}(k)$ for each candidate viewpoint $k\in\mathcal{C}$. It is trained with binary cross entropy over the candidate-view logits:
\begin{equation}
\mathcal{L}_{\pi_{\text{view}}}
=
\frac{1}{|\mathcal{C}|}
\sum_{k\in\mathcal{C}}
\mathrm{BCEWithLogits}
\left(
\hat{\ell}_{\text{view}}(k),
\tau_{\text{view}}(k)
\right).
\end{equation}

\paragraph{Push policy.}
The push policy is trained analogously using candidate-level push gains $g_{\text{push}}(p)$. The gains are normalized over valid push candidates and converted into sigmoid soft targets:
\begin{equation}
\tilde{g}_{\text{push}}(p)
=
\frac{
g_{\text{push}}(p)-\mu_{\mathcal{P}_t}(g_{\text{push}})
}{
\sigma_{\mathcal{P}_t}(g_{\text{push}})+\varepsilon
},
\qquad
\tau_{\text{push}}(p)
=
\sigma\!\left(\frac{\tilde{g}_{\text{push}}(p)}{T}\right).
\end{equation}
% With valid-candidate mask $M(p)$, the masked binary cross entropy loss is

The push policy outputs one logit $\hat{\ell}_{\text{push}}(p)$ for each candidate push $p\in\mathcal{P}_t$. With valid-candidate mask $M(p)$, the masked binary cross entropy loss is

\begin{equation}
\mathcal{L}_{\pi_{\text{push}}}
=
\frac{1}{\sum_p M(p)+\varepsilon}
\sum_{p\in\mathcal{P}_t}
M(p)\,
\mathrm{BCEWithLogits}
\left(
\hat{\ell}_{\text{push}}(p),
\tau_{\text{push}}(p)
\right).
\end{equation}
This objective trains the policy to learn relative candidate quality within the current push set, rather than a calibrated absolute gain across different scenes.

\paragraph{Belief map estimator.}
The belief map estimator is trained with an occupancy loss on the Beta mean and a semantic cross-entropy loss. Let
\begin{equation}
\mu_t^{\text{occ}}(v)
=
\frac{\alpha_t(v)}{\alpha_t(v)+\beta_t(v)}
\end{equation}
denote the predicted occupancy probability. We optimize
\begin{equation}
\mathcal{L}_{f_{\text{scene}}}
=
\frac{
\sum_v M_t(v)
\left(
\mu_t^{\text{occ}}(v)-y^{\text{occ}}(v)
\right)^2
}{
\sum_v M_t(v)+\varepsilon
}
+
w_{\text{sem}}
\frac{
\sum_v M_t(v)\,
\mathrm{CE}(\hat{s}_t(v),y^{\text{sem}}(v))
}{
\sum_v M_t(v)+\varepsilon
},
\end{equation}
where $M_t$ is the valid supervision mask.
For the scratch checkpoint used by the main model, we set $w_{\text{sem}}=0.1$. For class-OOD finetuning, we start from this pretrained $n_{\text{sem}}=15$ checkpoint and use the finetuning default $w_{\text{sem}}=1.0$. The semantic prediction is fed back at the next step as a detached softmax distribution.

\subsection{Training Procedure}
\label{app:hyperparams}

All modules are trained in simulation and selected by validation performance. DiVE and the two mode-specific policies are trained on low- and high-occlusion scenes. Extreme-occlusion scenes are not used for training and are reserved for held-out evaluation. For the class-OOD experiment, only the belief map estimator is lightly finetuned for the newly introduced semantic classes; DiVE and the action-selection policies remain frozen.

For DiVE, we use teacher forcing at the beginning of training and linearly anneal the teacher-forcing probability to zero, so the model gradually transitions to self-feeding its own visibility predictions. The push-resolvable branch self-feeds throughout training because its target is defined as a marginal action-induced visibility-change map rather than a temporally accumulated view-frequency state.

For policy learning, candidate gains are not regressed directly. Instead, each candidate set is normalized independently using $z$-score normalization and converted into sigmoid soft targets with temperature $T=1.0$. This trains the policies to predict relative candidate quality within the current decision state, which is the quantity needed at inference.

For the belief map estimator, we first train a scratch checkpoint on low- and high-occlusion YCB scenes with $n_{\text{sem}}=15$ and $w_{\text{sem}}=0.1$. For the class-OOD setting, we finetune only $f_{\text{scene}}$ from this checkpoint on unseen-v1 data using the reserved semantic slots, while keeping DiVE and both action-selection policies frozen. In both stages, the previous semantic belief is fed back as a detached softmax distribution, which stabilizes recurrent belief updates over long action sequences. The best checkpoint is selected using validation reconstruction quality, with occupancy IoU used as the primary selection criterion.

\section{Experimental Details}
\label{app:experimental_details}

\subsection{Baselines}
\label{app:baselines}

We compare DiVE with three reconstruction baselines and one visibility-decomposition ablation. All methods are evaluated under the same voxel grid, candidate viewpoint set $\mathcal{C}$, candidate push set $\mathcal{P}_t$, object categories, action budget, and evaluation protocol. Unless otherwise stated, all learned components are trained on the same simulation dataset described in Sec.~\ref{app:data_generation}. This ensures that the comparison isolates the effect of action selection and visibility decomposition rather than differences in scene distribution, sensing geometry, or evaluation budget.

\paragraph{CNABU.}
CNABU is a belief-gain planning baseline based on the map-space belief prediction framework of Marques et al.~\citep{Marques2025}. It represents the scene belief as a metric-semantic map and uses learned belief updates to predict the effect of sensing and manipulation actions. At each decision step, CNABU evaluates candidate view and push actions by estimating their predicted post-action beliefs and scoring them with volumetric information gain. The action with the highest predicted gain is then selected.

In our experiments, CNABU is trained on our dataset and evaluated with the same action budget, candidate viewpoint set $\mathcal{C}$, and candidate push set $\mathcal{P}_t$ as DiVE. Unlike DiVE, CNABU does not use decomposed view-resolvable and push-resolvable visibility maps. Instead, it relies on candidate-wise post-action belief prediction and information-gain evaluation. This makes CNABU a strong belief-gain planning baseline, but also computationally more expensive at test time because action selection requires evaluating the predicted gain of candidate actions online.

\paragraph{View-Only.}
View-Only is an ablation of DiVE that removes physical interaction from the action space. At every step, the planner is forced to select the viewing mode and chooses the next camera pose using $\pi_{\text{view}}$ over the candidate viewpoint set $\mathcal{C}$. The push policy $\pi_{\text{push}}$ is never invoked, and no swept volume is passed to the belief update. All other components, including the view-resolvable visibility update and the scene belief estimator, are kept unchanged.

This baseline isolates the role of pushing. If hidden regions can be recovered by sensing alone, View-Only should approach DiVE. However, in high- and extreme-occlusion scenes, some regions are not visible from any accessible viewpoint until the scene geometry is physically changed. In those cases, View-Only is expected to plateau because it can continue moving the camera but cannot remove occluders or expose physically hidden objects.

\paragraph{Random.}
Random is included as a lower-bound exploration baseline. It uniformly samples the next viewpoint from $\mathcal{C}$ without using DiVE, CNABU, belief gain, visibility maps, or any learned action-selection policy. The resulting observation is fused into the scene representation using a classical occupancy update. Because Random does not reason about which regions are unresolved or which actions are likely to reveal hidden space, it measures the benefit of informed active exploration over uninformed viewpoint sampling.

We use Random only as a weak exploration reference, not as a direct architectural ablation of DiVE. In particular, poor Random performance should not be interpreted as evidence for any single component of DiVE; rather, it indicates the gap between active visibility-aware planning and uninformed sensing.

\paragraph{Unified visibility ablation.}
Unified visibility is a visibility-decomposition ablation that removes the separation between view-resolvable and push-resolvable signals. Instead of predicting two modality-specific maps, the model predicts a single visibility map $\phi_{\text{unified}}$ that combines the remaining view-resolvable visibility change and the push-resolvable visibility change.

For this ablation, we construct the unified supervision target as
\begin{equation}
\phi^{*}_{\text{unified},t}(v)
=
\phi^{*}_{\text{view-rem},t}(v)
+
\phi^{*}_{\text{push},t}(v),
\end{equation}
where the remaining view-resolvable component is defined as
\begin{equation}
\phi^{*}_{\text{view-rem},t}(v)
=
\left|
\frac{1}{|\mathcal{C}|}
\sum_{k\in\mathcal{C}} m_k(v)
-
\frac{1}{|\mathcal{K}_t|}
\sum_{k\in\mathcal{K}_t} m_k(v)
\right|.
\end{equation}
Here, the first term is the marginal visibility of voxel $v$ over all candidate viewpoints, and the second term is the visibility already accumulated from the visited viewpoints. Their absolute difference measures the remaining discrepancy between the current view-resolvable visibility state and the full candidate-view marginal visibility. The push component $\phi^{*}_{\text{push},t}$ is the same target used to train the push-resolvable head in DiVE.

The same training data, voxel grid, candidate viewpoint set, and candidate push set are used as in DiVE, but the two visibility heads are replaced by a single output head. This ablation tests whether the performance gain comes merely from predicting a visibility-like map, or from preserving the action modality responsible for making each region visible. A unified map can indicate that a region is visibility-relevant, but it does not specify whether that region is better resolved by camera motion or by physical interaction.

\subsection{Action-Space and Planner Sensitivity Details}
\label{app:sensitivity}

We vary one action-space parameter at a time while keeping all other components fixed. Table~\ref{tab:action_space_sensitivity} reports the change in final occupancy IoU, in percentage points, relative to the default configuration of 98 viewpoints, 20 push candidates, a 15~cm push length, and a $\pm30^\circ$ direction range. Performance changes remain small across the tested alternatives, indicating that the gains are not tied to a single candidate-set parameterization.

We also compare the gain-based mode selector with fixed schedules that push every 3, 5, or 7 actions (Table~\ref{tab:planner_sensitivity}). All fixed schedules reduce final IoU on high-occlusion scenes. Increasing the post-push warm-up from the default of two accumulated views to three, four, or five views also provides no improvement, supporting the two-view constraint used in the main experiments.

\subsection{Test Scene Generation}
\label{app:test_scene_generation}

We evaluate simulation performance on held-out Isaac Sim test scenes constructed from YCB objects. Each scene is captured from the same set of 98 candidate viewpoints used by the planner, with RGB-D observations and ground-truth voxel-level occupancy and semantic annotations. All test scenes are generated from random seeds that are disjoint from the training and validation sets.

\begin{figure}[t]
\centering
\begin{subfigure}[t]{0.32\linewidth}
\centering
\includegraphics[width=\linewidth]{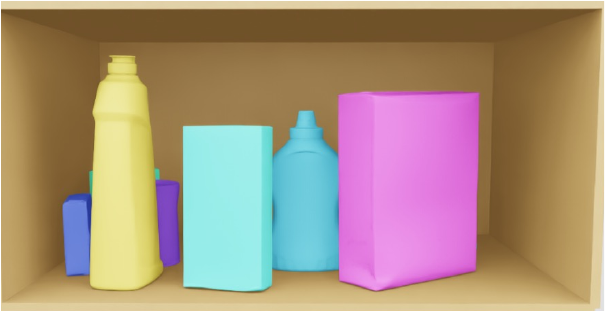}
\caption{Low occlusion}
\label{fig:low_sim_scene}
\end{subfigure}
\hfill
\begin{subfigure}[t]{0.32\linewidth}
\centering
\includegraphics[width=\linewidth]{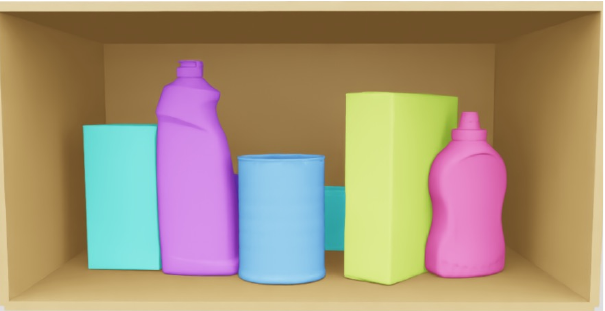}
\caption{High occlusion}
\label{fig:high_sim_scene}
\end{subfigure}
\hfill
\begin{subfigure}[t]{0.32\linewidth}
\centering
\includegraphics[width=\linewidth]{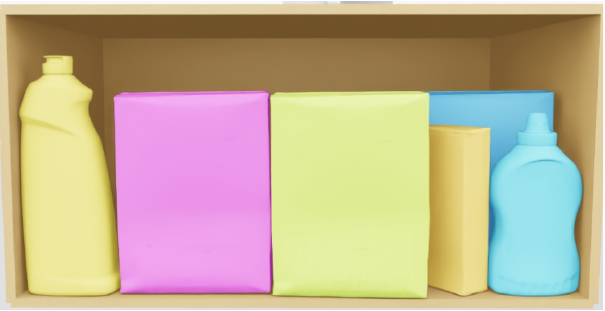}
\caption{Extreme occlusion}
\label{fig:extreme_sim_scene}
\end{subfigure}
\caption{Example simulation scenes from the three in-distribution occlusion regimes. Low-occlusion scenes contain sparse object arrangements where most objects are visible from at least one viewpoint. High-occlusion scenes use denser placements with stronger mutual occlusion. Extreme-occlusion scenes further increase packing density, causing several objects to be fully hidden from all candidate viewpoints before interaction.}
\label{fig:sim_occlusion_examples}
\end{figure}

The in-distribution test scenes are designed to vary occlusion difficulty while keeping the object category set and object geometry fixed. We use three occlusion regimes: low, high, and extreme occlusion. Figure~\ref{fig:sim_occlusion_examples} shows representative simulation scenes from these three occlusion regimes.

\paragraph{Low-occlusion scenes.}
Low-occlusion scenes are generated using the same YCB category set and a placement regime close to the training distribution. Each scene contains 11 active objects placed with collision-free sampling and sufficient inter-object spacing. As a result, most object instances are visible from at least one of the candidate viewpoints. On average, only 1.08 objects per scene are fully hidden from all 98 views, with at most 3 fully hidden objects in a scene. This split measures performance under the standard visibility regime in which most hidden regions can be recovered by viewpoint changes.

\paragraph{High-occlusion scenes.}
High-occlusion scenes use the same object pool and capture protocol, but the placement sampler is tightened to create denser configurations with more frequent contact, stacking, and mutual occlusion. Each scene again contains 11 active objects, but the average number of objects fully hidden from all candidate views increases to 1.44, with a maximum of 5. This split tests whether the model remains effective when larger portions of the scene are not visible from individual viewpoints and the belief estimator must reason over more occluded volume.

\paragraph{Extreme-occlusion scenes.}
Extreme-occlusion scenes form the main stress-test regime. They are generated with even denser object placement and contain 12 active objects per scene, one more than the low- and high-occlusion splits. These scenes frequently contain strong image-plane overlap and objects placed behind other objects. On average, 4.28 objects per scene are fully hidden from all 98 candidate views, with 4 to 5 fully hidden objects in every scene. This split evaluates performance under severe clutter, where successful reconstruction requires stronger amodal completion and occlusion-aware belief estimation.

Together, these three in-distribution splits form an occlusion-difficulty ladder: low-occlusion scenes evaluate standard active reconstruction, high-occlusion scenes increase the amount of hidden geometry, and extreme-occlusion scenes test whether the model can maintain useful beliefs when several objects are completely unobservable before interaction.

\subsection{Out-of-Distribution Evaluation}
\label{app:ood_evaluation}

We further evaluate generalization using two out-of-distribution test splits. Both splits keep the placement regime close to the low-occlusion setting so that the geometric difficulty remains comparable to the in-distribution low-occlusion split. This allows us to attribute performance changes mainly to object-level distribution shift rather than to increased occlusion.

\paragraph{Class-OOD scenes.}
The class-OOD split uses the same low-occlusion placement regime, but replaces the training object categories with held-out YCB categories that do not appear in the training set. Specifically, we use two unseen categories, 024\_bowl and 025\_mug, which are excluded from the in-distribution YCB object set in Table~\ref{tab:ycb_objects}. These categories introduce semantic and geometric structures not observed during training, including a concave bowl shape and a mug handle. Each scene contains 9 active objects, reflecting the smaller held-out category pool. The average number of objects fully hidden from all candidate views is 1.08. Since the occlusion level is close to the in-distribution low-occlusion split, this evaluation isolates generalization to unseen object categories rather than increased occlusion difficulty.

\paragraph{Shape-OOD scenes.}
The shape-OOD split uses the same in-distribution category set and the same low-occlusion placement regime, but changes the object geometry by applying anisotropic per-axis scaling to each object instance. Specifically, the scale factor for each object axis is sampled uniformly from $[0.8, 1.2]$, producing geometric variations that are not seen during training while preserving the original category identity. Each scene contains 11 active objects. The average number of fully hidden objects is 1.12, keeping the occlusion difficulty close to the in-distribution low-occlusion setting. This split evaluates whether the model depends on memorized object shapes or can generalize its visibility and belief predictions to geometric variations of known categories.

The class-OOD and shape-OOD splits are complementary. The former changes semantic identity while keeping geometry relatively standard, whereas the latter keeps category identity fixed while changing geometry. Comparing them helps separate degradation caused by category-level priors from degradation caused by shape-level priors.

\subsection{Real World}
\label{app:real_world_details}

\subsubsection{Real-World Setup}
\label{app:real_world_setup}

\begin{figure}[t]
\centering
\begin{subfigure}[t]{0.36\linewidth}
\centering
\includegraphics[width=\linewidth]{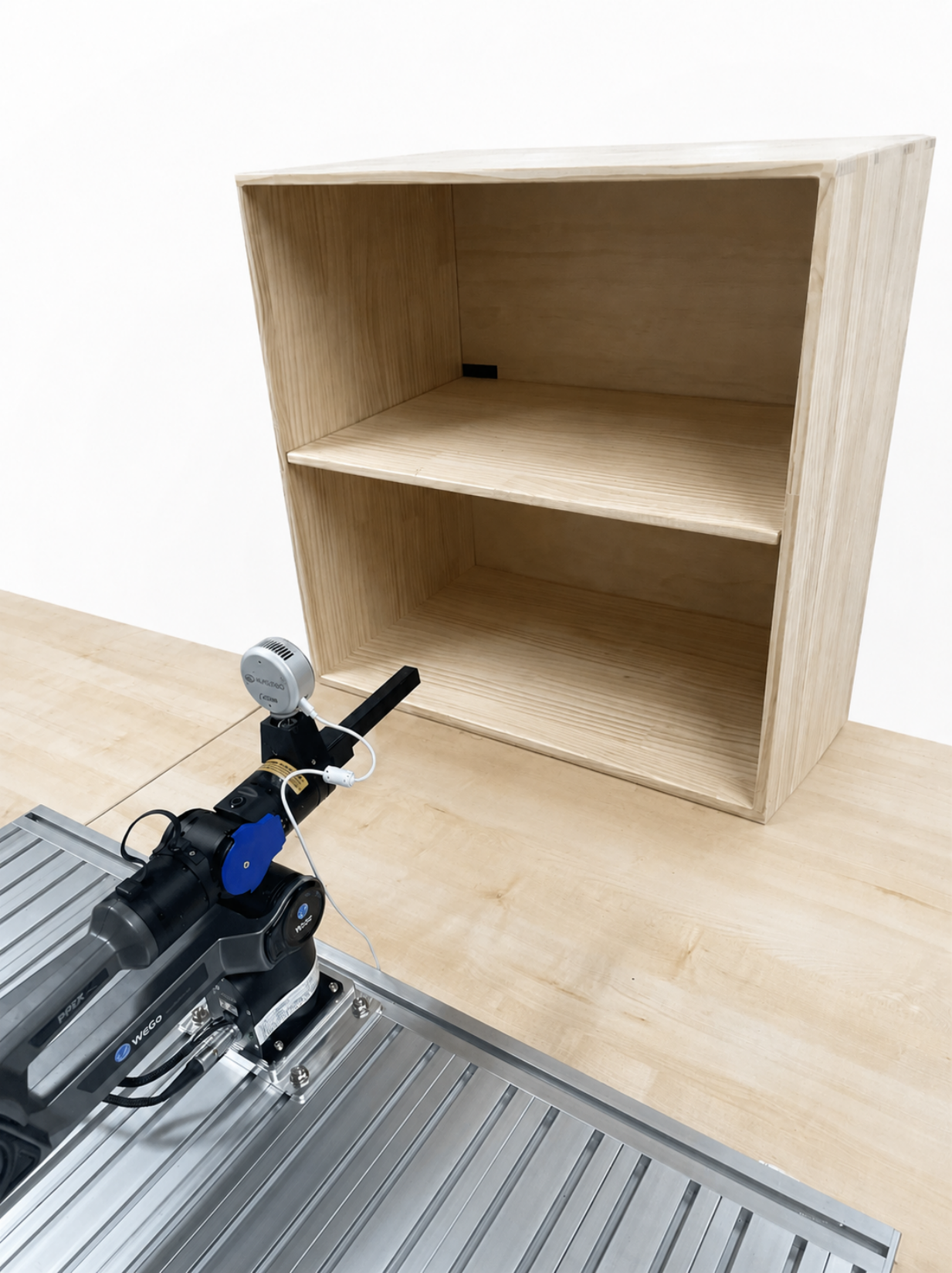}
\caption{Real-world robot and shelf setup.}
\label{fig:real_world_setup_photo}
\end{subfigure}
\hfill
\begin{subfigure}[t]{0.60\linewidth}
\centering
\includegraphics[width=\linewidth]{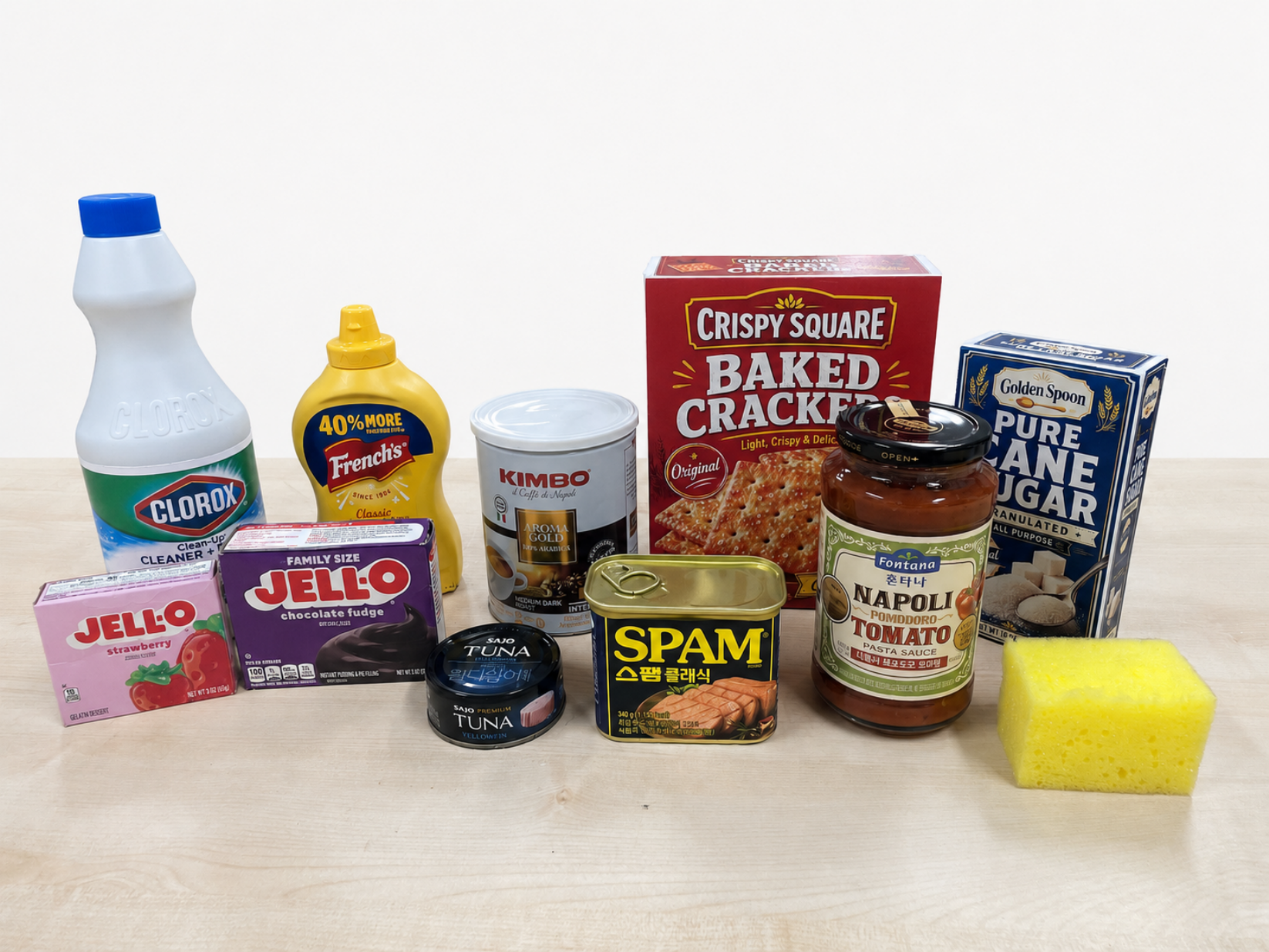}
\caption{Physical objects used in real-world scenes.}
\label{fig:real_world_objects_photo}
\end{subfigure}
\caption{Real-world experimental setup. The robot observes a cluttered shelf using an end-effector-mounted RGB-D camera and performs viewpoint changes or pushing actions selected by the planner. The object set consists of physical instances corresponding to the YCB-category objects used in simulation, but with real-world differences in shape, material, texture, and sensing noise.}
\label{fig:real_world_setup}
\end{figure}

We deploy DiVE zero-shot on a real robot shelf setup. The physical setup uses a Piper AgileX manipulator positioned in front of a shelf with geometry comparable to the simulation environment. An RGB-D camera is mounted on the robot end-effector and provides partial observations of the shelf scene from selected viewpoints. The robot executes the same type of view-and-push exploration policy as in simulation: at each step, it either moves the camera to a selected viewpoint or executes a pushing action to alter the occlusion structure. Figure~\ref{fig:real_world_setup} shows the real-world robot setup and the physical object set used in the experiments.

\paragraph{Robot and camera.}
The robot platform is a Piper AgileX manipulator. An Intel RealSense LiDAR Camera L515 is mounted near the end-effector and provides RGB-D observations from the selected viewpoints. Camera poses are calibrated with respect to the robot base frame, allowing observations from different viewpoints to be integrated into the global voxel grid. The shelf geometry is chosen to be comparable to the simulation workspace, so the same voxelized reconstruction pipeline and view-and-push action structure can be used without retraining the planner.

\paragraph{Object set.}
The real-world scenes are constructed using physical objects that correspond to the YCB categories used in simulation. Although the category labels are shared with simulation, the real objects differ in exact size, shape, texture, reflectance, and material. This makes the real-world experiment a zero-shot transfer test for the visibility-based planner under sensing noise and appearance shift.

\paragraph{Scene construction.}
We construct 10 real-world cluttered shelf scenes. Each scene contains multiple YCB-category objects arranged to create partial occlusions. Some objects are visible in the initial observation, while others are partially or fully occluded by front objects. The initial arrangement is manually created but not optimized for any method.

\paragraph{Observation and reconstruction.}
At each step, the RGB-D observation is converted into the same voxel grid representation used in simulation. The belief map estimator updates the global scene belief from the current observation. After a pushing action, the swept volume of the push is also passed to the belief update so that the estimator can account for the region physically affected by the interaction.

\paragraph{Execution protocol.}
Each real-world episode uses a fixed budget of 20 actions. We compare DiVE with View-Only under the same initial scene configurations. DiVE may choose either viewpoint changes or pushes, whereas View-Only is restricted to camera motion. No real-world finetuning is performed; the deployed policy is transferred zero-shot from simulation.

\paragraph{Real-world evaluation.}
For real-world episodes, we evaluate reconstruction at the object level rather than with dense voxel-level ground truth, since precise voxel annotation is unreliable under real sensing noise, calibration error, and manipulation-induced object motion. After each episode, we manually inspect the scene and record whether each object is correctly discovered, geometrically found but semantically misclassified, or missed. For initially occluded objects, we additionally record whether they become discovered during exploration.

\begin{figure}[t]
\centering
\includegraphics[width=0.75\linewidth]{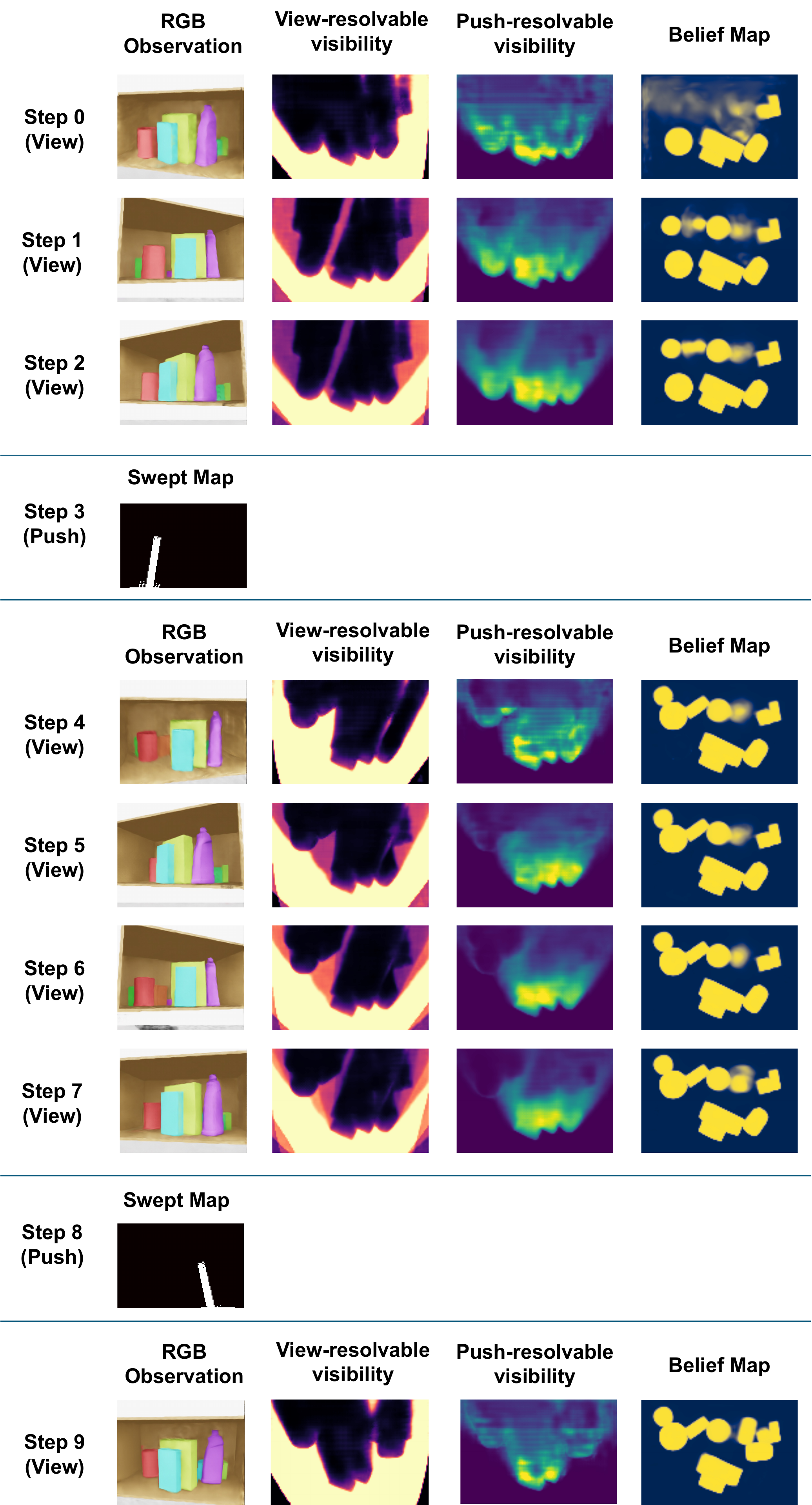}
\caption{
Qualitative internal rollout of DiVE in simulation.
For view steps, each row shows the RGB observation, the top-down projection of the view-resolvable visibility map $\phi_{\text{view}}$, the push-resolvable visibility map $\phi_{\text{push}}$, and the occupancy belief map.
For push steps, we show the swept map of the selected push action, which marks the region physically affected by interaction.
The sequence illustrates how DiVE alternates between viewpoint changes and pushing: view actions refine visible and view-resolvable regions, while push actions modify the scene and update the belief through the swept region.
}
\label{fig:dive_internal_rollout_sim}
\end{figure}

\section{Result Details}
\label{app:result_details}

\subsection{Qualitative Rollout Analysis}
\label{app:qualitative_rollout}

Figure~\ref{fig:dive_internal_rollout_sim} visualizes a representative internal rollout of DiVE in simulation. For view steps, we show the RGB observation, the view-resolvable visibility map $\phi_{\text{view}}$, the push-resolvable visibility map $\phi_{\text{push}}$, and the resulting occupancy belief map. For push steps, we visualize the swept volume of the selected push action, which is provided to the belief update after interaction.

The rollout illustrates how DiVE alternates between sensing and interaction during exploration. In the early steps, the planner selects view actions while the scene still contains regions that can be resolved from additional viewpoints. After several observations, the planner selects a push action, and the corresponding swept map indicates the region physically affected by the interaction. The belief map is then updated after the push, allowing subsequent view actions to refine the reconstructed scene. A second push is selected later in the episode, showing that DiVE can repeatedly return to interaction when the remaining scene structure requires physical rearrangement.

This qualitative example complements the main quantitative results by showing how decomposed visibility maps and push-conditioned belief updates evolve over an exploration episode. While this visualization does not show the candidate-level policy scores, it demonstrates that the selected action sequence produces meaningful changes in the belief map, especially after push actions.

\subsection{Hierarchical Action Selection Analysis}
\label{app:policy_output_comparison}

\begin{figure}[t]
    \centering
    \includegraphics[width=0.99\linewidth]{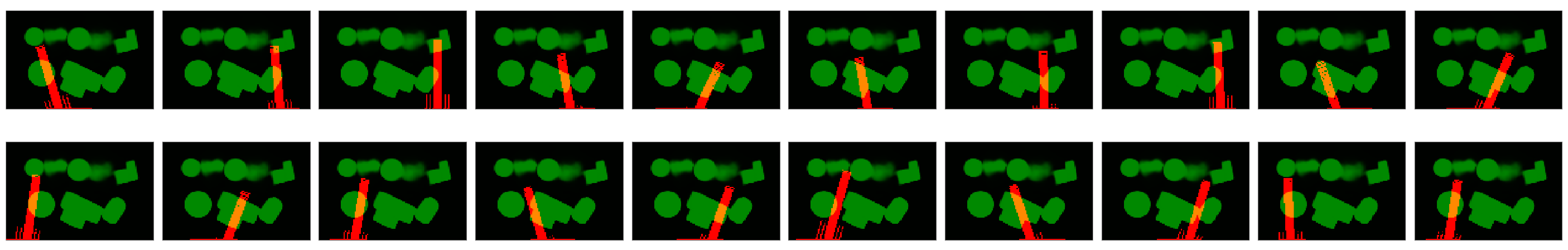}
    \caption{
    Visualization of the 20 push candidates considered at the decision step analyzed in Fig.~\ref{fig:policy_output_comparison}.
    Each panel shows the swept map $\Omega_p$ of one candidate push over the current scene, indicating the region that would be physically affected by executing that push.
    The push policy $\pi_{\text{push}}$ ranks these candidates within the selected pushing mode and chooses the candidate predicted to produce the largest push-induced visibility gain.
    }
    \label{fig:push_candidates}
\end{figure}

To further analyze action selection, we visualize the push-candidate set and the resulting policy decision at the same simulation step. Figure~\ref{fig:push_candidates} first shows the 20 push candidates considered at this decision step. Each candidate is represented by its swept map, which marks the voxels that would be physically affected if the corresponding push were executed. This visualization clarifies the input space of the push policy: $\pi_{\text{push}}$ does not directly optimize over a continuous interaction space, but ranks a finite set of sampled push candidates whose swept regions overlap different parts of the current scene.

\begin{figure}[t]
    \centering
    \includegraphics[width=0.99\linewidth]{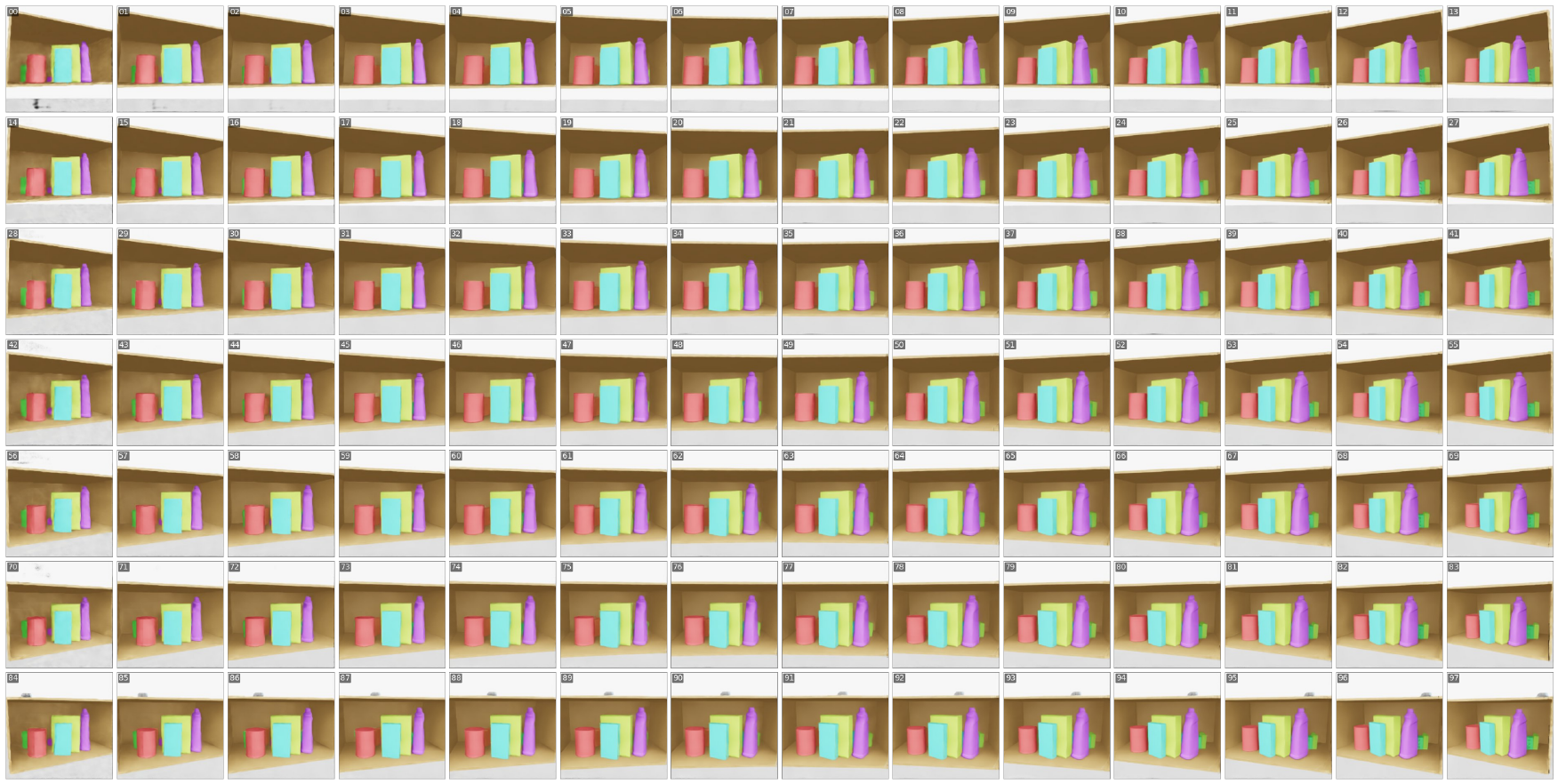}
    \caption{
    Visualization of the 98 candidate viewpoints considered at the decision step analyzed in Fig.~\ref{fig:policy_output_comparison}.
    Each panel shows the RGB observation that would be obtained from one candidate viewpoint of the current scene.
    The view policy $\pi_{\text{view}}$ ranks these candidates within the viewing mode and selects the viewpoint predicted to yield the largest view-resolvable visibility gain.
    }
    \label{fig:view_candidates}
\end{figure}

\begin{figure}[t]
    \centering
    \includegraphics[width=1.0\linewidth]{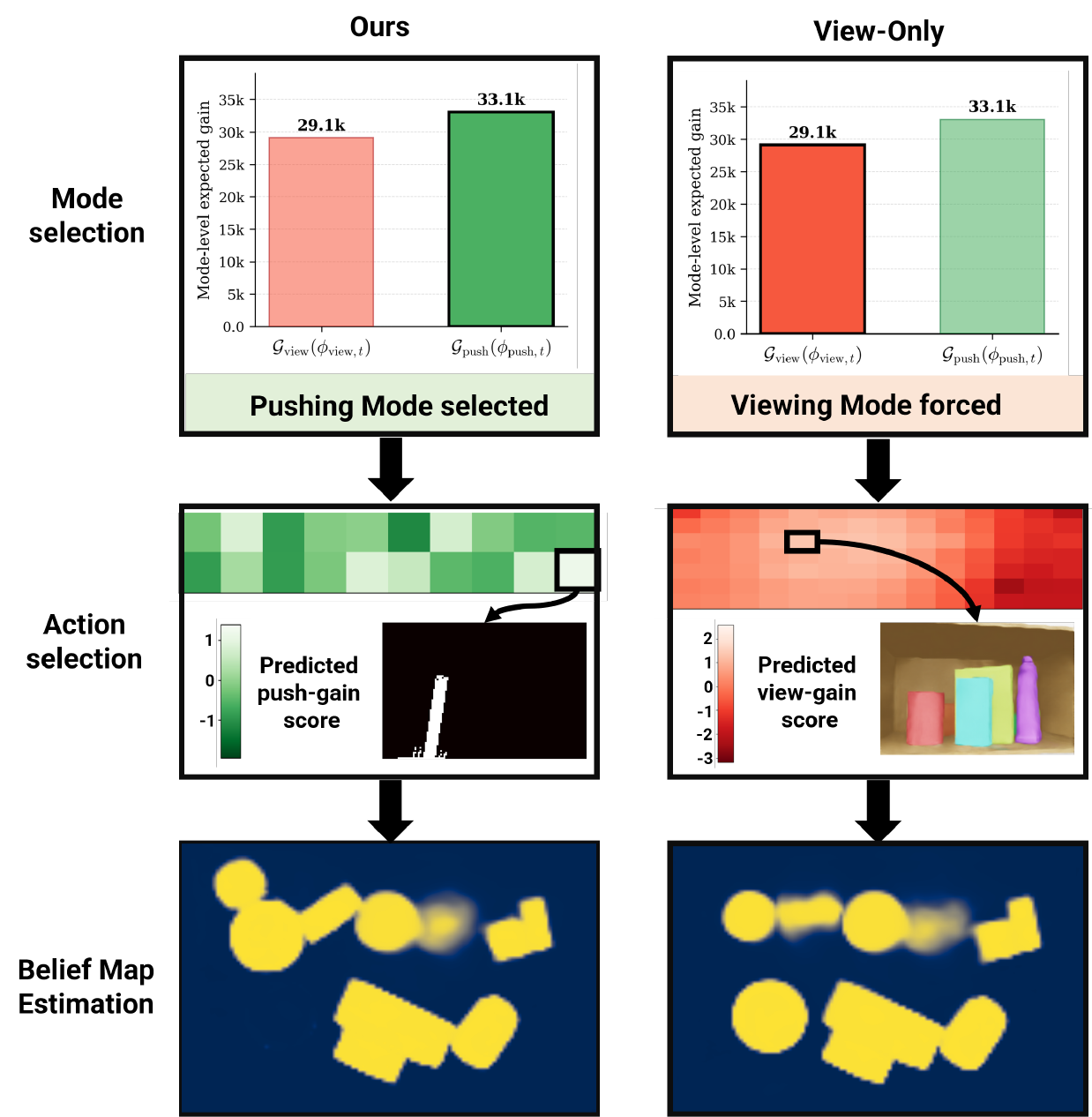}
    \caption{
    % Candidate-level action-selection comparison at the same decision step in simulation.
    % Left: DiVE (ours). The mode-level expected-gain comparison favors pushing, so the planner selects the pushing mode, ranks the push candidates shown in Fig.~\ref{fig:push_candidates}, and executes the selected push, whose swept region is shown.
    % Right: View-Only. Under the same scene state, the baseline is restricted to viewpoint changes and is therefore forced to select an additional view, despite the larger push-resolvable gain.
    % The updated belief maps show that DiVE's push action reveals and updates previously unresolved regions more effectively than an additional viewpoint alone.
    Candidate-level action-selection comparison at the same decision step in simulation.
    Left: DiVE (ours). The mode-level expected-gain comparison favors pushing, so the planner selects the pushing mode, ranks the push candidates shown in Fig.~\ref{fig:push_candidates}, and executes the selected push, whose swept region is shown.
    Right: View-Only. Under the same scene state, the baseline is restricted to viewpoint changes and is therefore forced to select an additional view from the candidate viewpoints visualized in Fig.~\ref{fig:view_candidates}, despite the larger push-resolvable gain.
    The updated belief maps show that DiVE's push action reveals and updates previously unresolved regions more effectively than an additional viewpoint alone.
    }
    \label{fig:policy_output_comparison}
\end{figure}

Figure~\ref{fig:policy_output_comparison} then compares DiVE and the View-Only baseline at this same decision step. While Fig.~\ref{fig:dive_internal_rollout_sim} visualizes how the internal maps evolve over time, this comparison focuses on how the planner converts the visibility signals into a specific action choice. In the DiVE panel, the push-candidate heatmap summarizes the policy scores assigned to the candidates shown in Fig.~\ref{fig:push_candidates}, and the selected candidate is visualized through its swept region.

At this scene state, the mode-level expected-gain comparison indicates that pushing is more informative than viewing, i.e., $G_{\text{push}}(\phi_{\text{push},t}) > G_{\text{view}}(\phi_{\text{view},t})$. DiVE therefore selects the pushing mode, ranks the candidate pushes using $\pi_{\text{push}}$, and executes the selected action whose swept region is shown in Fig.~\ref{fig:policy_output_comparison}. In contrast, the View-Only baseline is restricted to viewpoint changes and is therefore forced to select another view, even though the push-resolvable gain is larger at the same state.

The resulting belief maps highlight the practical consequence of this difference. DiVE's push action updates the unresolved region more effectively by physically altering the occlusion structure, whereas the additional viewpoint selected by View-Only provides more limited improvement. This qualitative comparison complements the quantitative results in the main paper by showing that DiVE not only decides when interaction is needed, but also benefits from selecting the appropriate action mode before ranking candidates within that mode.

\subsection{Additional-Primitive Feasibility}
\label{app:additional_primitive}

To examine whether the proposed representation can accommodate an
interaction primitive beyond pushing, we train an additional
grasp-resolvable visibility head.
Its supervision is constructed from the marginal visibility change
produced by removing a grasped object from the scene.
Unlike the view- and push-resolvable heads, this head therefore learns
to localize regions that could become visible through object removal.

\begin{figure}[t]
    \centering
    \includegraphics[width=0.72\linewidth]
    {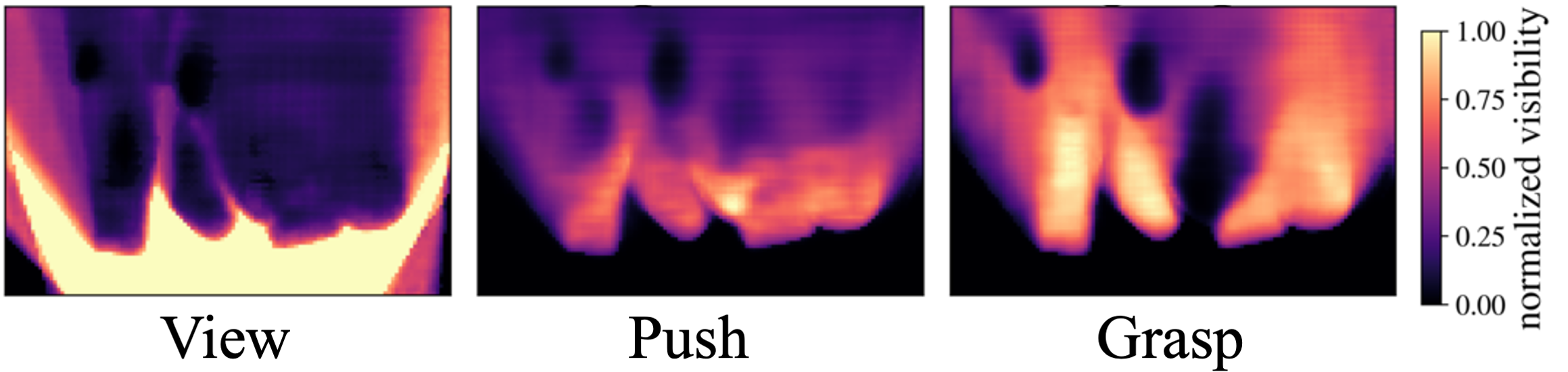}
    \vspace{-5pt}
    \caption{\textbf{Primitive-specific visibility maps.}
    Predicted view-, push-, and grasp-resolvable visibility maps for the
    same scene. The grasp-resolvable head is trained using the marginal
    visibility change caused by removing a grasped object.}
    \label{fig:additional_primitive_maps}
    \vspace{-5pt}
\end{figure}

Figure~\ref{fig:additional_primitive_maps} shows the predicted
view-, push-, and grasp-resolvable visibility maps.
The grasp-resolvable response is spatially localized around regions
whose visibility changes after object removal, indicating that the
modality-conditioned representation can be instantiated for an
additional manipulation primitive. This experiment is intended only as a representation-level proof of
concept.
It does not evaluate closed-loop grasp selection, joint
view--push--grasp exploration episodes, grasp execution success, or
improvements on a downstream manipulation task.

\subsection{Unified Visibility Ablation Analysis}
\label{app:unified_visibility_analysis}

We further analyze the unified visibility ablation by examining whether the predicted map provides a reliable candidate-ranking signal for push actions. For each push candidate $p$ with swept volume $\Omega_p$, we compute a swept-volume score by summing the predicted visibility response over the voxels affected by the push:
\begin{equation}
s_{\text{unified}}(p)
=
\sum_{v \in \Omega_p}
\phi_{\text{unified}}(v),
\qquad
s_{\text{push}}(p)
=
\sum_{v \in \Omega_p}
\phi_{\text{push}}(v).
\end{equation}
Here, $s_{\text{unified}}(p)$ is computed from the unified visibility map, whereas $s_{\text{push}}(p)$ is computed from the decomposed push-resolvable map. We compare both scores against the true push-induced visibility gain over the candidate push set.

\begin{figure}[t]
    \centering
    \includegraphics[width=0.75\linewidth]{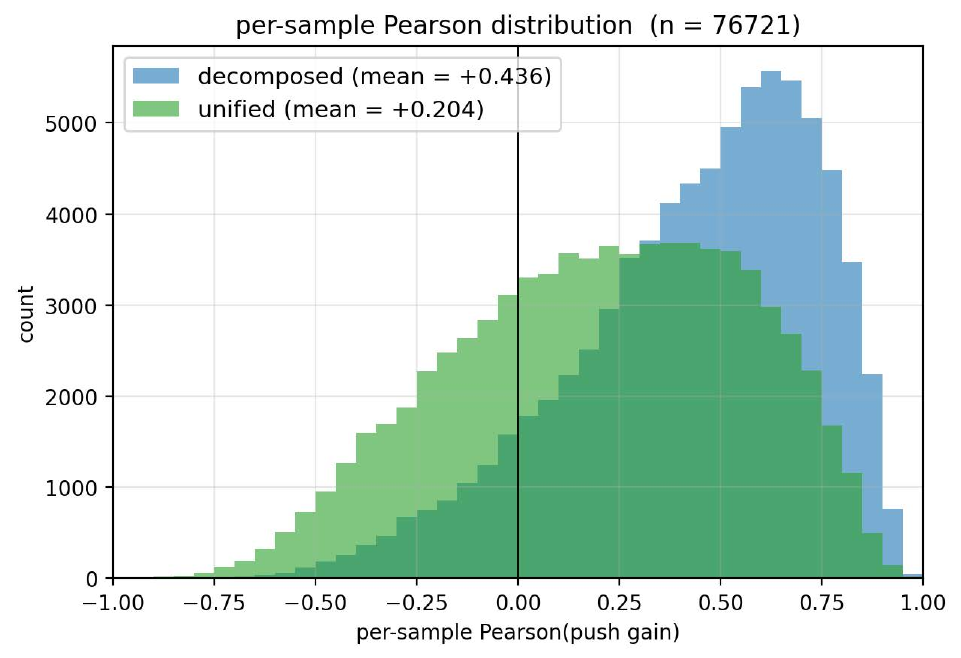}
    \caption{
    Distribution of per-sample Pearson correlations between swept-volume map scores and true push-induced visibility gain.
    The decomposed push-resolvable map $\phi_{\text{push}}$ yields a distribution shifted toward stronger positive correlations, showing that it provides a more reliable push-candidate ranking signal.
    The unified visibility map produces a broader distribution with more weak or negative correlations, indicating that mixing view-resolvable and push-resolvable signals dilutes the push-specific cue required for interaction selection.
    }
    \label{fig:unified_visibility_distribution}
\end{figure}

Figure~\ref{fig:unified_visibility_distribution} shows the distribution of per-sample Pearson correlations between the swept-volume scores and the true push gain. The decomposed push-resolvable map produces a distribution that is shifted toward positive correlation, indicating that high responses in $\phi_{\text{push}}$ tend to overlap with push candidates that actually produce large visibility changes. In contrast, the unified map produces a broader distribution with substantially more weak or negative correlations. This means that the unified map often assigns high response to regions that are visibility-relevant in a general sense, but not necessarily useful for selecting a push action.

This result highlights the role of visibility decomposition in candidate-level action selection. A single unified map can represent that a region remains visibility-relevant, but it does not preserve which action modality is responsible for resolving that region. Consequently, swept-volume scores computed from the unified map can be contaminated by view-resolvable regions that should be handled by camera motion rather than physical interaction. The decomposed push-resolvable map avoids this ambiguity by retaining a push-specific signal, making candidate push ranking more aligned with the actual effect of interaction.

\end{document}